\documentclass[twoside,11pt]{article}

\usepackage{jmlr2e}
\usepackage{amsmath}
\usepackage{dsfont}
\usepackage{booktabs}
\usepackage{algorithm,algpseudocode}
\usepackage{mathtools}
\usepackage{centernot}

\newcommand{\CI}{\mathrel{\perp\mspace{-10mu}\perp}}

\firstpageno{1}

\begin{document}

\title{Feature-Level Domain Adaptation}


\author{\name Wouter M. Kouw \email w.m.kouw@tudelft.nl \\
       \name Laurens J.P. van der Maaten \email l.j.p.vandermaaten@tudelft.nl \\
       \addr Department of Intelligent Systems\\
       Delft University of Technology\\
       Mekelweg 4, 2628 CD Delft, The Netherlands \\
       \AND
       \name Jesse H. Krijthe \email jkrijthe@gmail.com \\
       \addr Department of Intelligent Systems\\
       Delft University of Technology\\
       Mekelweg 4, 2628 CD Delft, The Netherlands \\
       \addr Department of Molecular Epidemiology \\
       Leiden University Medical Center \\
       Einthovenweg 20, 2333 ZC Leiden, The Netherlands \\
       \AND
       \name Marco Loog \email m.loog@tudelft.nl \\
       \addr Department of Intelligent Systems\\
       Delft University of Technology\\
       Mekelweg 4, 2628 CD Delft, the Netherlands \\
       \addr The Image Group\\
       University of Copenhagen \\
       Universitetsparken 5, DK-2100 Copenhagen, Denmark \\
       \AND
}

\editor{}

\maketitle

\begin{abstract}%
Domain adaptation is the supervised learning setting in which the training and test data are sampled from different distributions: training data is sampled from a source domain, whilst test data is sampled from a target domain. This paper proposes and studies an approach, called feature-level domain adaptation ({\sc flda}), that models the dependence between the two domains by means of a feature-level transfer model that is trained to describe the transfer from source to target domain. Subsequently, we train a domain-adapted classifier by minimizing the expected loss under the resulting transfer model. For linear classifiers and a large family of loss functions and transfer models, this expected loss can be comp	uted or approximated analytically, and minimized efficiently. Our empirical evaluation of {\sc flda} focuses on problems comprising binary and count data in which the transfer can be naturally modeled via a dropout distribution, which allows the classifier to adapt to differences in the marginal probability of features in the source and the target domain. Our experiments on several real-world problems show that {\sc flda} performs on par with state-of-the-art domain-adaptation techniques.
\end{abstract}


\section{Introduction}

Domain adaptation is an important research topic in machine learning and pattern recognition that has applications in, among others, speech recognition (\citealp{leggetter1995maximum}), medical image processing (\citealp{van2013transfer}), computer vision (\citealp{saenko2010adapting}), natural language processing (\citealp{peddinti2011domain}), and bioinformatics (\citealp{borgwardt2006integrating}). Domain adaptation deals with supervised-learning settings in which the common assumption that the training and the test observations stem from the same distribution is dropped. This learning setting may arise, for instance, when the training data is collected with a different measurement device than the test data, or when a model that is trained on one data source is deployed on data that comes from another data source. This creates a learning setting in which the training set contains samples from one distribution (the so-called source domain), whilst the test set constitutes samples from another distribution (the target domain). In domain adaptation, one generally assumes a transductive learning setting: that is, it is assumed that the unlabeled test data are available to us at training time and that the main goal is to predict their labels as well as possible.

The goal of domain-adaptation approaches is to exploit information on the dissimilarity between the source and target domains that can be extracted from the available data in order to make more accurate predictions on samples from the target domain. To this end, many domain adaptation approaches construct a {\it sample-level transfer model} that assigns weights to observations from the source domain in order the make the source distribution more similar to the target distribution (\citealp{shimodaira2000improving, cortes2011domain, gretton2009covariate, huang2007correcting, cortes2008sample}). In contrast to such sample-level reweighing approaches, in this work, we develop a {\it feature-level transfer model} that describes the shift between the target and the source domain for each feature individually. Such a feature-level approach may have advantages in certain problems: for instance, when one trains a natural language processing model on news articles (the source domain) and applies it to Twitter data (the target domain), the marginal distribution of some of the words or n-grams (the features) is likely to vary between target and source domain. This shift in the marginal distribution of the features cannot be modeled well by sample-level transfer models, but it can be modeled very naturally by a feature-level transfer model.

Our feature-level transfer model takes the form of a conditional distribution that, conditioned on the training data, produces a probability density of the target data. In other words, our model of the target domain thus comprises a convolution of the empirical source distribution and the transfer model. The parameters of the transfer model are estimated by maximizing the likelihood of the target data under the model of the target domain. Subsequently, our classifier is trained as to minimize the expected value of the classification loss under the target-domain model. We show empirically that when the true domain shift can be modeled by the transfer model, under certain assumptions, our domain-adapted classifier converges to a classifier trained on the true target distribution. Our feature-level approach to domain adaptation is general in that it allows the user to choose a transfer model from a relatively large family of probability distributions. This allows practitioners to incorporate domain knowledge on the type of domain shift in their models. In the experimental section of this paper, we focus on a particular type of transfer distribution that is well-suited for problems in which the features are binary or count data (as often encountered in natural language processing), but the approach we describe is more generally applicable. In addition to experiments on artificial data, we present experiments on several real-world domain adaptation problems, which show that our feature-level approach performs on par with the current state-of-the-art in domain adaptation.

The outline of the remainder of this paper is as follows. In Section 2, we give an overview of related prior work on domain adaptation. Section 3 presents our feature-level domain adaptation ({\sc flda}) approach. In Section 4, we present our empirical evaluation of feature-level domain adaptation. Section 5 concludes the paper with a discussion of our results.

\section{Related Work}
Current approaches to domain adaptation can be divided into one of three main types. The first type constitutes {\it importance weighting} approaches that aim to reweigh samples from the source distribution in an attempt to match the target distribution as well as possible. The second type are {\it sample transformation} approaches that aim to transform samples from the source distribution in order to make them more similar to samples from the target distribution. The third type are {\it feature augmentation} approaches that aim to extract features that are shared across domains. Our feature-level domain adaptation ({\sc flda}) approach is an example of a sample-transformation approach.

\paragraph{Importance weighting.} Importance-weighting approaches assign a weight to each source sample in such a way as to make the reweighted version of the source distribution as similar to the target distribution as possible (\citealp{shimodaira2000improving, cortes2011domain, gretton2009covariate, huang2007correcting, cortes2008sample, gong2013connecting, baktashmotlagh2014domain}). If the class posteriors are identical in both domains (that is, the covariate-shift assumption holds) and the importance weights are unbiased estimates of the ratio of the target density to the source density, then the importance-weighted classifier converges to the classifier that would have been learned on the target data if labels for that data were available (\citealp{shimodaira2000improving}). Despite their theoretic appeal, importance-weighting approaches generally do not to perform very well when the dataset is small, or when there is little ''overlap'' between the source and target domain. In such scenarios, only a very small set of samples from the source domain is assigned a large weight. As a result, the effective size of the training set on which the classifier is trained is very small, as a result of which a poor classification model may be obtained. In contrast to importance-weighting approaches, our approach performs a {\it feature-level} reweighing. Specifically, {\sc flda} assigns a data-dependent weight to each of the features that represents how informative this feature is in the target domain. This approach effectively uses all the data in the source domain, as a result of which it does not suffer from the small sample size problem.

\paragraph{Sample transformation.} Sample-transformation approaches learn transformations of the source data and target data that try to make the source distribution more similar to the target distribution (\citealp{pan2011domain, gopalan2011domain, baktashmotlagh2013unsupervised, gong2012geodesic, blitzer2006domain, dinh2013fidos, fernando2013unsupervised, shao2014generalized, blitzer2011domain}). Most sample-transformation approaches learn global (non)linear transformations that map source and target data points into the same, shared feature space in such a way as to maximize the overlap between the transformed source data and the transformed target data (\citealp{gopalan2011domain,gong2012geodesic,fernando2013unsupervised,pan2011domain, baktashmotlagh2013unsupervised}). Approaches that learn a shared subspace in which both the source and the target data are embedded often minimize the maximum mean discrepancy (MMD) between the transformed source data and the transformed target data (\citealp{pan2011domain, baktashmotlagh2013unsupervised}). When used in combination with a universal kernel, the MMD criterion is zero when all the moments of the (transformed) source and target distribution are identical. Most methods minimize the MMD subject to constraints that help to avoid trivial solutions (such as collapsing all data onto the same point) via some kind of spectral analysis. An alternative to the MMD is the subspace disagreement measure (SDM) of \cite{gong2012geodesic}, which measures the discrepancy of the angles between the principal components of the transformed source data and the transformed target data. Most current sample-transformation approaches work well for ``global'' domain shifts such as translations or rotations in the feature space, but they are less effective for when the domain shift is ''local'' in the sense that it strongly nonlinear. Similar limitations apply to the {\sc flda} approach we explore, but it differs in that (1) our transfer model does not learn a subspace but operates in the original feature space and (2) the measure it minimizes to model the transfer is different, namely, the negative log-likelihood of the target data under the transferred source distribution.

\paragraph{Feature augmentation.} Several domain-adaptation approaches extend the source data and the target data with additional features that are similar in both domains (\citealp{li2014learning, blitzer2006domain}). Specifically, the approach by \cite{blitzer2006domain} tries to induce correspondences between the features in both domains by identifying so-called pivot features that appear frequently in both domains but that behave differently in each domain; SVD is applied on the resulting pivot features to obtain a low-dimensional, real-valued feature representation that is used to augment the original features. This approach works well for natural language processing problems due to the natural presence of correspondences between features, e.g. words that signal each other. The approach of \cite{blitzer2006domain} is related to the many of the instantiations of {\sc flda} that we consider in this paper, but it is different in the sense that we only use information on differences in feature presence between the source and the target domain to reweigh those features (that is, we do not augment the feature representation). Moreover, the formulation of {\sc flda} is more general, and can be extended through a relatively large family of transfer models.

\section{Feature-Level Domain Adaptation}
Suppose we wish to train a sentiment classifier for reviews, and we have a dataset with book reviews and associated sentiment labels (positive or negative review) available. After having trained a linear classifier on word-count representations of the book reviews, we wish to deploy it to predict the sentiment of kitchen appliance reviews. This leaves us with a domain-adaptation problem on which the classifier trained on book reviews will likely not work very well: the classifier will assign large positive weights to, for instance, words such as ''interesting'' and ''insightful'' as these suggest positive book reviews and will be assigned large positive weights by a linear classifier, but these words hardly ever appear in reviews of kitchen appliances. As a result, a classifier trained on the book reviews in a naive way may perform poorly on kitchen appliance reviews. Since the target domain data (the kitchen appliance reviews) are available at training time, a natural approach to resolving this problem may be to down-weight features corresponding to words that do not appear in the target reviews, for instance, by applying a high level of dropout (\citealp{hinton2012improving}) to the corresponding features in the source data when training the classifier. The use of dropout mimics the target domain scenario in which the ''interesting'' and ''insightful'' features are hardly ever observed during the training of the classifier, and prevents that these features are assigned large positive weights during training. Feature-level domain adaptation {\sc flda} aims to formalize this idea in a two-stage approach that (1) fits a probabilistic sample transformation model that aims to model the transfer between source and target domain and (2) trains a classifier by minimizing the risk of the source data under the transfer model. 

In the first stage, {\sc flda} models the transfer between the source and the target domain: the transfer model is a data-dependent distribution that models the likelihood of target data conditioned on observed source data. Examples of such transfer models may be a dropout distribution that assigns a likelihood of $1-\theta$ to the observed feature value in the source data and a likelihood of $\theta$ to a feature value of $0$, or a Parzen density estimator in which the mean of each kernel is shifted by a particular value. The parameters of the transfer distribution are learned by maximizing the likelihood of target data under the transfer distribution (conditioned on the source data). In the second stage, we train a linear classifier to minimize the expected value of a classification loss under the transfer distribution. For quadratic and exponential loss functions, this expected value and its gradient can be analytically derived whenever the transfer distribution factorizes over features and is in the natural exponential family; for logistic and hinge losses, practical upper bounds and approximations can be derived (\citealp{van2013learning,wager2013dropout,chen2014dropout}). 

In the experimental evaluation of {\sc flda}, we focus on applying dropout transfer models to domain-adaptation problems involving binary and count features. These features frequently appear in, for instance, bag-of-words features in natural language processing (\citealp{blei2003latent}) or bag-of-visual-words features in computer vision (\citealp{jegou2012aggregating}). However, we note that {\sc flda} can be used in combination with a larger family of transfer models; in particular, the expected loss that is minimized in the second stage of {\sc flda} can be computed or approximated efficiently for any transfer model that factorizes over variables and that is in the natural exponential family.

\subsection{Notation}

We assume a domain adaptation setting in which we receive pairs of samples and labels from the {\it source} domain, $S = \{ ({\bf x}_{i}, y_{i}) \mid {\bf x}_{i} \sim p_{\mathcal{X}}, {\bf x}_i \in \mathbb{R}^{m}, y_{i} \in Y \}_{i = 1,\dots, N_{S}}$, at training time. Herein, the set $Y$ is assumed to be a set of discrete classes and $p$ refers to the probability distribution of its subscripted variable (${\cal X}$ for the source domain variable, ${\cal Z}$ for the target domain variable and ${\cal Y}$ for the class variable). At test time, we receive samples from the {\it target} domain, $T = \{{\bf z}_{j} \mid {\bf z}_{j} \sim p_{\mathcal{Z}}, {\bf z}_{j} \in \mathbb{R}^{m}\}_{j = 1,\dots, N_{T}}$ that need to be classified. Note that we assume samples ${\bf x}_{i}$ and ${\bf z}_{j}$ to lie in the same feature space $\mathbb{R}^{m}$, hence, we assume that $p_{\mathcal{X}}$ and $p_{\mathcal{Z}}$ are distributions over the same space. For brevity, we occasionally adopt the notation $\mathbf{X} = \left[\mathbf{x}_1, \dots, \mathbf{x}_{|S|}\right]$, $\mathbf{Z} = \left[\mathbf{z}_1, \dots, \mathbf{z}_{|T|}\right]$, and $\mathbf{y} = \left[y_1, \dots, y_{|S|}\right]$.


\subsection{Target risk}
We adopt an empirical risk minimization (ERM) framework for constructing our domain-adapted classifier. The ERM framework proposes a classification function $h: \mathbb{R}^{m} \rightarrow \mathbb{R}$ and assesses the quality of the hypothesis by comparing its predictions with the true labels on the empirical data using a loss function $L: Y \times \mathbb{R} \rightarrow \mathbb{R}_{0}^{+}$. The empirical loss is an estimate of the \emph{risk}, which is the expected value of the loss function under the data distribution. Below, we show that if the target domain carries no additional information about the label distribution, the risk of a model on the target domain is equivalent to the risk on the source domain under a particular transfer distribution.

We first note that the joint source data, target data and label distribution can be decomposed into two conditional distributions and one marginal source distribution; $p_{\mathcal{Y},\mathcal{Z},\mathcal{X}} = p_{\mathcal{Y}\mid \mathcal{Z},\mathcal{X}} \ p_{\mathcal{Z} \mid \mathcal{X}} \ p_{\mathcal{X}}$. The first conditional $p_{\mathcal{Y} \mid \mathcal{Z},\mathcal{X}}$ describes the label distribution given both source and target distribution. Next, we introduce our main assumption: the labels are conditionally independent of the target domain given the source domain ($\mathcal{Y} \CI \mathcal{Z} \mid \mathcal{X}$), which implies: $p_{\mathcal{Y}\mid \mathcal{Z},\mathcal{X}} = p_{\mathcal{Y}\mid \mathcal{X}}$. In other words, we assume that we can construct an optimal target classifier if (1) we have access to infinitely many labeled source samples---we know $p_{\mathcal{Y}\mid \mathcal{X}} \ p_{\mathcal{X}}$---and (2) we know the true domain transfer distribution $p_{\mathcal{Z} \mid \mathcal{X}}$. In this scenario, observing target labels does not provide us with any new information that can be used to improve the target classifier. 

To illustrate the implications of our assumption, imagine a problem in which the goal is to predict whether a review of a product is positive or negative. If people frequently use the word ''nice'' in positive reviews about electronics products (the source domain), then we assume the word ''nice'' is not predictive of negative reviews of kitchen appliances (the target domain). Under this assumption, learning a good predictive model for the target domain amounts to transferring the source domain to the target domain (that is, altering the marginal probability of observing the word ''nice'') and learning a good predictive model on the resulting transferred source domain. Admittedly, there are scenarios in which our assumption is invalid: if people like ''small'' electronics but dislike ''small'' cars, the assumption is violated and our domain-adaptation approach will likely work less well. We do note, however, that our assumption is less stringent than the covariate-shift assumption, which assumes that the posterior distribution over classes is identical in the source and the target domain (\emph{i.e.}, that $p_{\mathcal{Y} \mid \mathcal{X}} = p_{\mathcal{Y} \mid \mathcal{Z}}$): the covariate-shift assumption does not facilitate the use of a transfer distribution $p_{{\cal Z} \mid {\cal X}}$.

We start by rewriting the risk $R$ on the target domain as follows:
\begin{align}
R(h) =& \int_{{\cal Z}}\sum_{y \in Y} \ L(y,h({\bf z})) \ p_{{\cal Y},{\cal Z}}(y,{\bf z}) \ \mathrm{d}{\bf z}  & \nonumber \\
=& \int_{{\cal Z}}\sum_{y \in Y}\int_{{\cal X}} \ L(y,h({\bf z})) \ p_{{\cal Y}, {\cal Z}, {\cal X}}(y,{\bf z},{\bf x}) \ \mathrm{d}{\bf x} \ \mathrm{d}{\bf z} & \nonumber \\
 =& \int_{{\cal Z}}\sum_{y \in Y}\int_{{\cal X}} \ L(y,h({\bf z})) \ p_{{\cal Y} \mid {\cal Z}, {\cal X}}(y \mid {\bf z},{\bf x}) \ p_{{\cal Z} \mid {\cal X}}({\bf z} \mid {\bf x})  \ p_{{\cal X}}({\bf x}) \ \mathrm{d}{\bf x} \ \mathrm{d}{\bf z}. & \label{tr1}
\end{align}

Using the assumption introduced above, we can rewrite this expression as:
\begin{align}
R(h) =& \int_{{\cal Z}}\sum_{y \in Y}\int_{{\cal X}} \ L(y,h({\bf z})) \ p_{{\cal Y} \mid {\cal X}}(y \mid {\bf x}) \ p_{{\cal Z} \mid {\cal X}}({\bf z} \mid {\bf x})  \ p_{{\cal X}}({\bf x}) \ \mathrm{d}{\bf x} \ \mathrm{d}{\bf z} \nonumber \\
 =& \int_{{\cal Z}} \mathbb{E}_{{\cal Y},{\cal X}}\left[ L(y,h({\bf z})) \ p_{{\cal Z}|{\cal X}}({\bf z} \mid {\bf x}) \right] \ \mathrm{d}{\bf z}. \label{tr2}
\end{align}

Next, we replace the expectation in Equation \ref{tr2} by an empirical estimate on the source data:
\begin{align}
\hat{R}(h \mid S) =&\  \frac{1}{| S |} \int_{{\cal Z}} \sum_{({\bf{x}_i}, y_i) \in S} L(y_{i},h({\bf z})) \ p_{{\cal Z}\mid {\cal X}} ({\bf z} \mid {\bf x}={\bf x}_{i}) \ \mathrm{d}{\bf z} & \nonumber \\
=&\  \frac{1}{| S |} \sum_{({\bf{x}_i}, y_i) \in S} \mathbb{E}_{{\cal Z} \mid {\cal X}={\bf{x}_i}} \left[ L(y_{i},h({\bf z})) \right]. \label{tr4}
\end{align}

Feature-level domain adaptation ({\sc flda}) trains classifiers by constructing a parametric model of the transfer distribution $p_{{\cal Z} \mid {\cal X}}$ and, subsequently, minimizing the expected loss in Equation~\ref{tr4} \emph{on the source data} with respect to the parameters of the classifier. For linear classifiers, the expected loss in Equation~\ref{tr4} can be computed analytically for quadratic and exponential losses if the transfer distribution factorizes over dimensions and is in the natural exponential family; for the logistic and hinge losses, it can be upper-bounded or approximated efficiently under the same assumptions (\citealp{van2013learning, wager2013dropout,chen2014dropout}).

Note that no observed target samples ${\bf z}_{j}$ are involved Equation \ref{tr4}; the expectation is over the transfer model $p_{{\cal Z} \mid {\cal X}}$, conditioned on a particular sample ${\bf x}_i$. Whilst we do not use the target samples when training the final classifier, we do target data to estimate the parameters of the transfer model as described below.

\subsection{Transfer model}
\label{Transfer model}
The transfer distribution $p_{{\cal Z} \mid {\cal X}}$ describes the relation between the source and the target domain: given a source sample, it produces a distribution over corresponding target samples (that are assumed to have the same label as the source sample). The transfer distribution is modeled by selecting a parametric distribution and learning the parameters of this distribution from the source and target data (without looking at the source labels). Prior knowledge on the relation between source and target domain may be incorporated in the model via the choice of the (family of) distributions. For instance, if we know that the main variation between two domains is that particular words are frequently used in one domain (say, news articles) but infrequently in another domain (say, tweets), then we select a distribution that alters the relative frequency of words. 

Given a model of the transfer distribution $p_{{\cal Z} \mid {\cal X}}$ and a model of the source distribution $p_{\cal X}$, we can work out the marginal distribution over the target domain as:
\begin{align}
	q_{{\cal Z}}({\bf z} \mid \theta, \eta) = \int_{{\cal X}} \ p_{{\cal Z} \mid {\cal X}}({\bf z} \mid {\bf x}, \theta) \ p_{{\cal X}}({\bf x} \mid \eta) \ \mathrm{d}{\bf x}, \label{transferred}
\end{align}
where $\theta$ represents the parameters of the transfer model, and $\eta$ the parameters of the source model. We learn these parameters separately: first, we learn $\eta$ by maximizing the likelihood of the source data under the model $p_{{\cal X}}({\bf x} \mid \eta)$ and, subsequently, we learn $\theta$ by maximizing the likelihood of the target data under the compound model $q_{{\cal Z}}({\bf z} \mid \theta, \eta)$. Hence, we first estimate the value of $\eta$ by solving:
\begin{equation}
	\hat{\eta} = \underset{\eta}{\arg \max} \sum_{{\bf x}_i \in T} \log p_{{\cal X}}({\bf x}_{i}\mid \eta).
\end{equation}
Subsequently, we estimate the value of $\theta$ by solving:
\begin{align}
	\hat{\theta} = \underset{\theta}{\arg \max} \sum_{{\bf z}_j \in T} \log q_{{\cal Z}}({\bf z}_{j}\mid \theta, \hat{\eta}). \label{max_tr_params}
\end{align}

In this paper, we focus primarily on domain-adaptation problems involving binary and count features. In such problems, we wish to encode a changes in the marginal likelihood of observing non-zero values in the transfer model. To this send, we employ a dropout distribution as transfer model that can model domain-shifts in which a feature occurs less often in the target domain than in the source domain. Learning a {\sc flda} model with a dropout transfer model has the effect of strongly regularizing weights on features that occur infrequently in the target domain.

\subsubsection{Dropout transfer} 
To define our transfer model for binary or count features, we first set up a model that describes the likelihood of observing non-zero features in the source data. This model comprises a product of independent Bernoulli distributions:
\begin{align}
p_{{\cal X}}({\bf x}_i \mid \eta) =& \ \prod_{d=1}^{m} \ \eta_{d}^{\mathds{1}_{x_{id} \neq 0}} \ (1-\eta_{d})^{1-\mathds{1}_{x_{id} \neq 0}}, \label{sourceB}
\end{align}
where $\mathds{1}$ is the indicator function and $\eta_{d}$ is the success probability (probability of non-zero values) of feature $d$. For this model, the maximum likelihood estimate of $\eta_{d}$ is simply the sample average: $\hat{\eta}_{d} = |S|^{-1}\sum_{{\bf x}_i \in S} \mathds{1}_{x_{id} \neq 0}$.

Next, we define a transfer model that describes how often a feature has a value of zero in the target domain when it has a non-zero value in the source domain. We assume an unbiased dropout distribution (\citealp{wager2013dropout, rostamizadeh2011learning}) that sets an observed feature in the source domain to zero in the target domain with probability $\theta_d$:
\begin{align}
p_{{\cal Z} \mid {\cal X}}(z_{-d} \mid x=x_{id}, \theta_{d}) &= \begin{cases}
\theta_{d} \quad & \text{if} \ z_{-d}=0\\	
1-\theta_{d} \quad \quad  &\text{if} \ z_{-d}=x_{id} \ / (1-\theta_{d}),
\end{cases} & \label{dropout}
\end{align}
where $\forall d: 0 \leq \theta_d \leq 1$, the subscript of $z_{-d}$ denotes the $d$-th feature for any target sample, and where the outcome of not {\it dropping out} is scaled by a factor $1/(1-\theta_d)$ in order to center the dropout distribution on the particular source sample. We assume the transfer distribution factorizes over features to obtain: $p_{{\cal Z} \mid {\cal X}}({\bf z} \mid {\bf x}={\bf x}_{i} , \theta) = \prod_{d}^{m} p_{{\cal Z} \mid {\cal X}}(z_{-d} \mid x_{-d}=x_{id}, \theta_{d})$. The equation above defines a transfer distribution for every source sample. We obtain our final transfer model by sharing the parameters $\theta$ between all transfer distributions and averaging over all source samples.

To compute the maximum likelihood estimate of $\theta$, the dropout transfer model from Equation \ref{dropout} and the source model from Equation \ref{sourceB} are plugged into Equation \ref{transferred} to obtain  (see Appendix A for details):
\begin{align}
	q_{{\cal Z}}({\bf z}\mid \theta, \eta) =& \prod_{d=1}^{m} \ \int_{{\cal X}} p_{{\cal Z} \mid {\cal X}}(z_{-d} \mid x_{-d}, \theta_{d}) \ p_{{\cal X}}(x_{-d} \mid \eta_{d}) \ \mathrm{d}x_{-d} \nonumber \\
	=& \ \prod_{d=1}^{m} \ \Big((1-\theta_{d}) \ \eta_{d}\Big)^{\mathds{1}_{{z}_{-d} \neq 0}} \ \Big(1-(1-\theta_{d}) \ \eta_{d}\Big)^{1-\mathds{1}_{{z}_{-d} \neq 0}}. \label{z_model}
\end{align}
Plugging this expression into Equation \ref{max_tr_params} and maximizing with respect to $\theta$, we obtain:
\begin{align}
	\hat{\theta}_{d} =& \max \left\{0, 1-\frac{\hat{\zeta}_{d}}{\hat{\eta}_{d}} \right\}, \nonumber 
\end{align}
where $\hat{\zeta}_{d}$ is the sample average of the dichotomized target samples, $|T|^{-1}\sum_{{\bf z}_j \in T} \mathds{1}_{z_{jd}\neq 0}$, and where $\hat{\eta}_{d}$ is the sample average of the dichotomized source samples, $|S|^{-1}\sum_{{\bf x}_i \in S} \mathds{1}_{x_{id} \neq 0}$.

We note that our particular choice for the transfer model cannot represent rate changes in the values of non-zero count features, such as whether a word is used on average 10 times in a document versus used on average only 3 times. The only variation that our dropout distribution captures is the variation in whether or not a feature occurs ($z_{-d}\neq 0$).

Because our dropout transfer model factorizes over features and is in the natural exponential family, the expectation in Equation \ref{tr4} can be analytically computed for quadratic and exponential loss functions. In particular, for a transfer distribution conditioned on source sample ${\bf x}_{i}$, the value of the expected value involves evaluation of the mean and variance:
\begin{align}
\mathbb{E}_{{\cal Z} \mid {\bf x}_{i}}\big[{\bf z} \big] =& \ {\bf x}_{i} \nonumber \\
\mathbb{V}_{{\cal Z} \mid {\bf x}_{i}}\big[{\bf z} \big] =& \ \text{diag}\left(\frac{\theta}{1-\theta}\right)  \circ {\bf x}_{i}{\bf x}_{i}^{\top} \nonumber,
\end{align}
where $\circ$ denotes the element-wise product of two matrices and we use the shorthand notation $\mathbb{E}_{{\cal Z} \mid {\cal X}={\bf x}_i} = \mathbb{E}_{{\cal Z} \mid {\bf x}_i}$. The variance is diagonal, because our dropout transfer model was defined to be independent across features. We will use these expressions below in our description of how to learn the parameters of the domain-adapted classifiers.

\subsection{Classification}
In order to perform classification with the risk formulation in Equation~\ref{tr4}, we need to select a loss function $L$. Popular choices for the loss function include the quadratic loss (used in least-squares classification), the exponential loss (used in boosting), the hinge loss (used in support vector machines) and the logistic loss (used in logistic regression). The formulation in (\ref{tr4}) has been studied before in the context of {\it dropout training} for the quadratic, exponential, and logistic loss by \cite{wager2013dropout,van2013learning}, and for hinge loss by \cite{chen2014dropout}. In this paper, we focus on the quadratic and logistic loss functions, but we note that the {\sc flda} approach can also be used in combination with exponential and hinge losses.

\subsubsection{Quadratic loss}
Assuming binary labels $Y = \{-1,+1\}$, a linear classifier with parameters $\mathbf{w}$, and a quadratic loss function $L$, the expectation in Equation \ref{tr4} can be expressed as:
\begin{align}
	\hat{R}({\bf w} \mid S) =& \sum_{({\bf x}_i, y_i) \in S} \mathbb{E}_{{\cal Z} \mid {\bf x}_i} \left[  (y_{i}-{\bf w}^{\top}{\bf z})^{2} \nonumber \right] \\
	=& \  {\bf y}^{\top}{\bf y} - 2 \  {\bf w}^{\top}\mathbb{E}_{{\cal Z} \mid \mathbf{X}}[{\bf Z}]{\bf y}^{\top}  + {\bf w}^{\top} \left( \mathbb{E}_{{\cal Z} \mid \mathbf{X}}[{\bf Z}] \mathbb{E}_{\cal{Z} \mid \mathbf{X}}[{\bf Z}]^\top + \mathbb{V}_{\cal{Z} \mid \mathbf{X}}[{\bf Z}] \right) {\bf w}, \label{qd_loss}
\end{align}
in which all feature vectors are appended with a value of $1$ to model the bias, and in which we denote $(m+1) \times |S|$ matrix of expectations $\mathbb{E}_{{\cal Z} \mid {\cal X} = \mathbf{X}}[{\bf Z}] = \begin{bmatrix}\mathbb{E}_{{\cal Z} \mid {\cal X} = {\bf x}_{1}}[{\bf z}], \dots, \ \mathbb{E}_{{\cal Z} \mid \cal{X} = {\bf x}_{|S|}}[{\bf z}] \end{bmatrix}$ and the $(m+1)\!\times\!(m+1)$ diagonal matrix of variances $\mathbb{V}_{\cal{Z} \mid {\cal X} =\mathbf{X}}[{\bf Z}] = \sum_{{\bf x}_i \in S} \mathbb{V}_{{\cal Z}\mid {\cal X}={\bf x}_{i}}[{\bf z}]$.

Deriving the gradient for this loss function and setting it to zero yields the closed-form solution for the classifier weights:
\begin{align}
{\bf w} =& \Big( \mathbb{E}_{{\cal Z} \mid \mathbf{X}}[{\bf Z}] \mathbb{E}_{{\cal Z} \mid \mathbf{X}}[{\bf Z}]^{\top} + \mathbb{V}_{{\cal Z} \mid \mathbf{X}}[{\bf Z}] \Big)^{-1} \mathbb{E}_{{\cal Z} \mid \mathbf{X}}[{\bf Z}]{\bf y}^{\top} \ . \label{flda-quad}
\end{align}
In the case of multi-class problem with $K$ classes ($Y = \{1,\dots,K\}$), $K$ predictors can be built in an one-vs-all fashion or $K(K-1)/2$ predictors in an one-vs-one fashion. 

The solution in Equation~\ref{flda-quad} is very similar to the solution of a standard ridge regression model (${\bf w} = \big( {\bf X} {\bf X}^{\top} + \lambda {\bf I}\big)^{-1}{\bf X}{\bf y}^{\top}$) trained on the source data: the main difference is that, in a ridge regressor, the regularization is independent of the data. By contrast, the regularization on the weights of the {\sc flda} solution is determined by the variance of the transfer model: hence, it is different for each dimension and it depends on the transfer from source to target domain. 

Algorithm \ref{flda-q} summarizes the training of binary {\sc flda} classifier that employs a quadratic loss and a dropout transfer model.
\begin{algorithm}[ht]
  \caption{Binary {\sc flda} with dropout transfer model and quadratic loss function.}\label{flda-q}
  \begin{algorithmic}
    \Procedure{flda-q}{$S, T$}
      \For{d=1,\dots , m}
      \State $\hat{\eta}_{d} = |S|^{-1} \sum_{{\bf x}_i \in S} \mathds{1}_{x_{id} \neq 0}$
      \State $\hat{\zeta}_{d} = |T|^{-1} \sum_{{\bf z}_j \in T} \mathds{1}_{z_{jd} \neq 0}$
        \State $\theta_{d} = \max \left\{ 0,1 - \hat{\zeta}_{d} \ / \ \hat{\eta}_{d} \right\}$ 
      \EndFor
	\State ${\bf w} = \left({\bf X}{\bf X}^{\top} + \text{diag}\left(\frac{\theta}{1-\theta}\right) \circ {\bf X}{\bf X}^{\top}\right)^{-1} {\bf X}{\bf y}^{\top}$ \Comment{$\circ$ Element-wise product}
      \State \textbf{return} \text{sign}(${\bf w}^{\top}{\bf Z}$) 
    \EndProcedure
  \end{algorithmic}
\end{algorithm}	

\subsubsection{Logistic loss}
Again, assuming binary labels $Y = \{-1,+1\}$, a linear classifier with parameters $\mathbf{w}$, the expectation in Equation \ref{tr4} for a logistic loss function $L$ can be expressed as:
\begin{align}
	\hat{R}(\mathbf{w} \mid S) =& \frac{1}{|S|} \sum_{({\bf x}_i, y_i) \in S} \mathbb{E}_{{\cal Z} \mid {\bf x}_i} \left[  -y_{i}{\bf w}^{\top}{\bf z} + \log\sum_{y' \in Y} \exp(y'{\bf w}^{\top}{\bf z})  \right] \nonumber \\
	=& \frac{1}{|S|} \sum_{({\bf x}_i, y_i) \in S} - y_{i} {\bf w}^{\top}\mathbb{E}_{{\cal Z} \mid {\bf x}_i}[{\bf z}] + \mathbb{E}_{{\cal Z} \mid {\bf x}_i}\left[\log \sum_{y'\in Y} \exp( y' {\bf w}^{\top}{\bf z})\right].\label{eq:logrisk}
\end{align}
This is a convex function in $\mathbf{w}$ because the log-sum-exp of an affine function is convex, and because the expected value of a convex function is convex. However, the expectation cannot be computed analytically. Following \cite{wager2013dropout}, we approximate the expectation of the log-partition function ($A({\bf w}^{\top}{\bf z})=\log \sum_{y'\in Y} \exp( y' {\bf w}^{\top}{\bf z})$) using a Taylor expansion around the value $a_i = \mathbf{w}^\top\mathbf{x}_i$:
\begin{align}
\mathbb{E}_{{\cal Z} \mid {\bf x}_i}\big[A({\bf w}^{\top}{\bf z})\big] &\approx A(a_i) + A'(a_i)(\mathbb{E}_{{\cal Z} \mid {\bf x}_i} [{\bf w}^{\top}{\bf z}] - a_i) + \frac{1}{2}A''(a_i)(\mathbb{E}_{{\cal Z} \mid {\bf x}_i} [{\bf w}^{\top}{\bf z}] - a_i)^{2} \nonumber \\
&= \mbox{const} + \sigma(-2\mathbf{w}^\top\mathbf{x}_i)\sigma(2\mathbf{w}^\top\mathbf{x}_i){\bf w}^{\top} \mathbb{V}_{{\cal Z} \mid {\bf x}_i}[{\bf z}]{\bf w},
\label{log_loss}
\end{align}
where $\sigma(x)= 1 / (1 +  \exp(-x))$ is the sigmoid function. In the Taylor approximation, the first-order term disappears because we defined our transfer model to be unbiased: $\mathbb{E}_{{\cal Z} \mid \mathbf{x}_i} [{\bf w}^{\top}{\bf z}] = \mathbf{w}^\top\mathbf{x}_i$. The approximation cannot be minimized in closed-form: we repeatedly take steps in the direction of its gradient in order to minimize the approximation of the risk in Equation~\ref{eq:logrisk}, as described in Algorithm \ref{flda-l} (see Appendix B for the gradient derivation). The algorithm can be readily extended to multi-class problems by replacing $\mathbf{w}$ by a $(m+1)\!\times\!K$ matrix and using an one-hot encoding for the labels (see Appendix C).
\begin{algorithm}[ht]
  \caption{{\sc flda} with dropout transfer model and logistic loss function.}\label{flda-l}
  \begin{algorithmic}
    \Procedure{flda-l}{$S, T$}
          \For{d=1,\dots,  m} 
      \State $\hat{\eta}_{d} = |S|^{-1} \sum_{{\bf x}_i \in S} \mathds{1}_{x_{id} \neq 0}$
      \State $\hat{\zeta}_{d} = |T|^{-1} \sum_{{\bf z}_j \in T} \mathds{1}_{z_{jd} \neq 0}$	
        \State $\theta_{d} = \max \left\{ 0,1 - \hat{\zeta}_{d} \ / \ \hat{\eta}_{d} \right\}$
      \EndFor
	      \State ${\bf w} = \underset{{\bf w'}}{\arg \min} - \sum_{({\bf x}_i, y_i) \in S} \left[ y_{i} {\bf w}^{\top}{\bf x}_i] \right]$\\
	      \hspace{4cm}  $+ {\bf w'}^{\top}\left(\sum_{({\bf x}_i, y_i) \in S} \left[\sigma \big(-2 \mathbf{w}^\top\mathbf{x}_i\big)\sigma \big(2 \mathbf{w}^\top\mathbf{x}_i\big) \text{diag}\left(\frac{ \theta}{1- \theta}\right) {\bf x}_i{\bf x}_i^{\top} \right]\right) {\bf w'}$ \\
      \State \textbf{return} \text{sign}(${\bf w}^{\top}{\bf Z}$)
    \EndProcedure
  \end{algorithmic}
\end{algorithm}

\section{Experiments}
In our experiments, we first study the empirical behavior of {\sc flda} on artificial data for which we know the true transfer distribution. Subsequently, we measure the performance of our method in a ''missing data at test time'' scenario, as well as on two image datasets and three text datasets with substantial domain transfer.

\subsection{Artificial data}
We first investigate the behavior of {\sc flda} on a problem in which the model assumptions are satisfied. We create such a problem setting by first sampling a source domain dataset from known class-conditional distributions. Subsequently, we construct a target domain dataset by sampling additional source data and transforming it using a pre-defined (dropout) transfer model. 

\subsubsection{Adaptation under correct model assumptions}
\begin{figure}[ht]
\centering
\includegraphics[width=.48\textwidth]{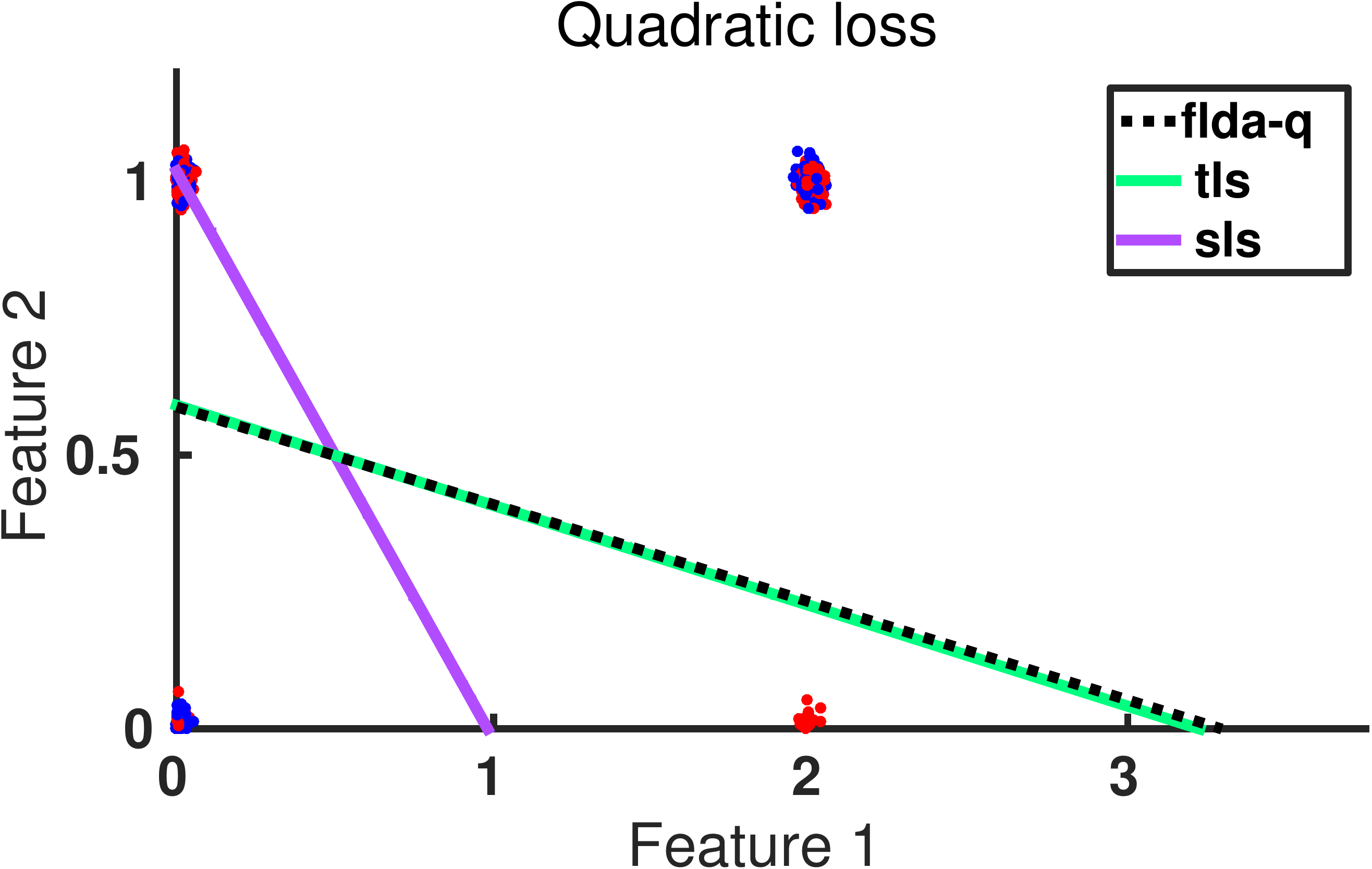} \hspace{5px}
\includegraphics[width=.48\textwidth]{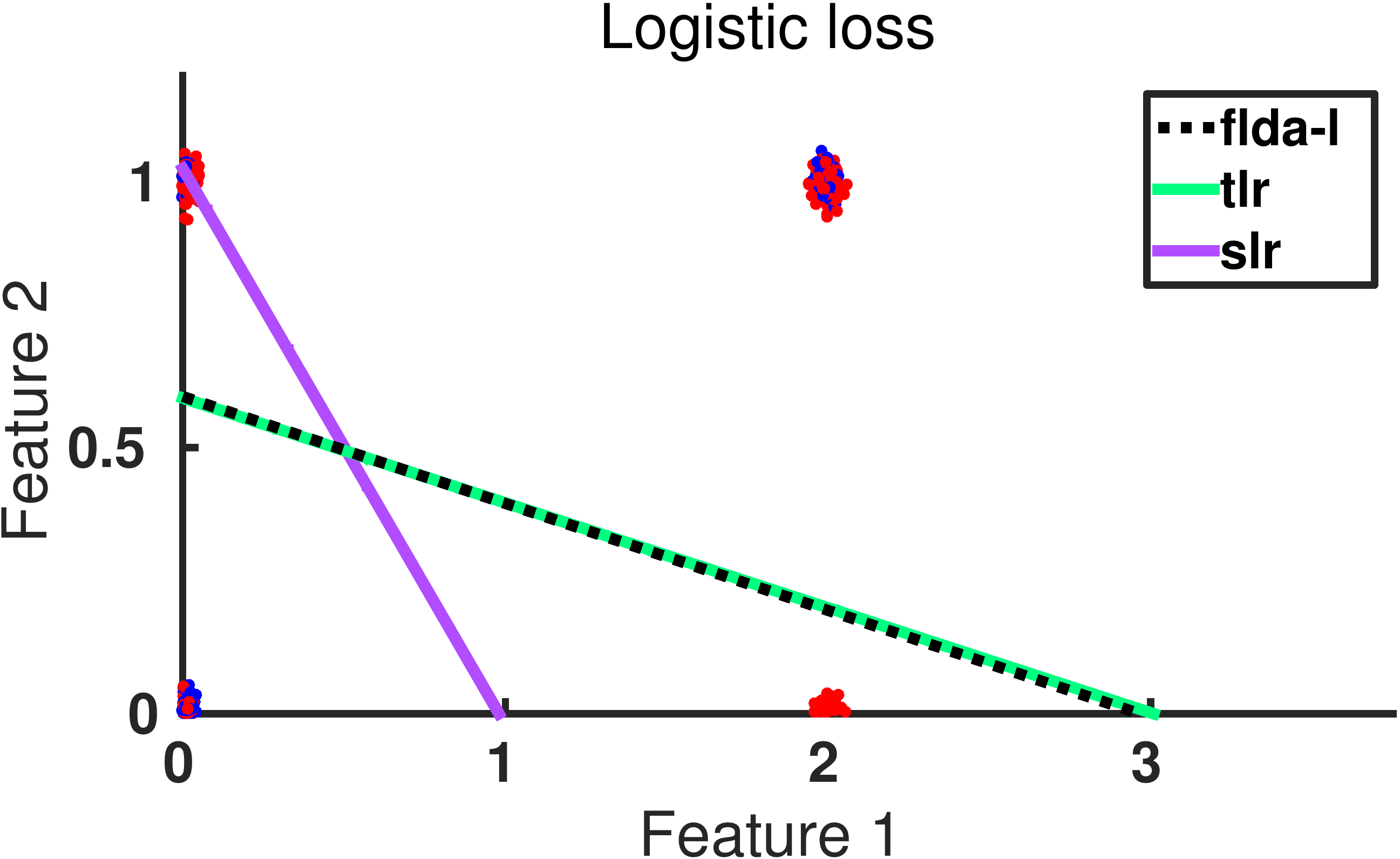}
\caption{Scatter plots of the target domain. The data is generated by bivariate Bernoulli class-conditional distributions and transformed using a dropout transfer. Red and blue dots show different classes. The lines are the decision boundaries found by the source classifier ({\sc s-lr}/{\sc s-ls}), the target classifier ({\sc t-lr}/{\sc t-ls}) and the adapted classifier (left {\sc flda-q}, right {\sc flda-l}). Note that the decision boundary of {\sc flda} lies on top of the decision boundary of {\sc t-lr}.}
\label{art1}
\end{figure}

We perform an experiment in which the domain-adapted classifier estimates the transfer model and trains the classifier on the source data; we evaluate the quality of the resulting classifier by comparing it to a classifier that was trained on the target data (that is, the classifier one would train if labels for the target data were available at training time). We consider a two-dimensional problem with binary features in which the data is generated by drawing $100,000$ samples from two bivariate Bernoulli distributions. The marginal distribution of both features is $\begin{bmatrix} 0.7 \ \  0.7\end{bmatrix}$ for class one and $\begin{bmatrix} 0.3 \ \  0.3\end{bmatrix}$ for class two. The source data is transformed to the target data using a dropout transfer model with parameters $\theta= \begin{bmatrix} 0.5 \ \  0\end{bmatrix}$. This means that 50\% of the values for feature 1 are set to 0 and the other values are scaled by $1/(1-0.5)$. For reference, two naive least-squares classifiers are trained, one on the source data ({\sc s-ls}) and one on the target data ({\sc t-ls}), and compared to {\sc flda-q}. {\sc s-ls} achieves a misclassification error of 0.400 while {\sc t-ls} and {\sc flda-q} achieve an error of 0.300. This experiment is repeated for the same classifiers but with logistic losses: a source logistic classifier ({\sc s-lr}), a target logistic classifier ({\sc t-lr}) and {\sc flda-l}. In this experiment, {\sc s-lr} achieves an error of 0.248 and {\sc t-lr} and {\sc flda-l} an error of 0.301. Figure \ref{art1} shows the decision boundaries for the quadratic loss classifiers on the left and the logistic loss classifiers on the right. The figure shows that for both loss functions, {\sc flda} has completely adapted to be equivalent to the target classifier in this artificial problem.

\begin{figure}[ht]
\centering
\includegraphics[width=.48\textwidth]{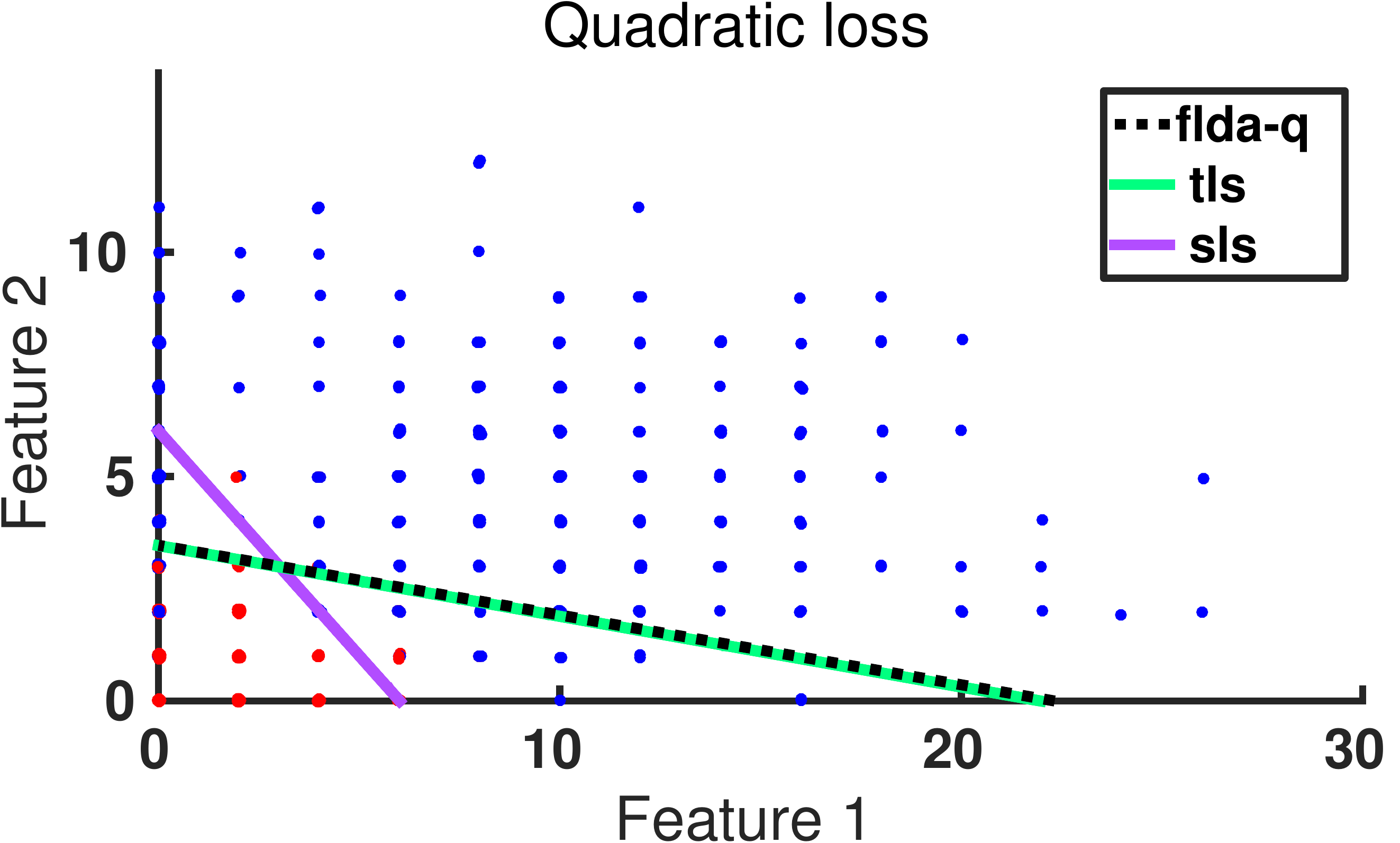} \hspace{5px}
\includegraphics[width=.48\textwidth]{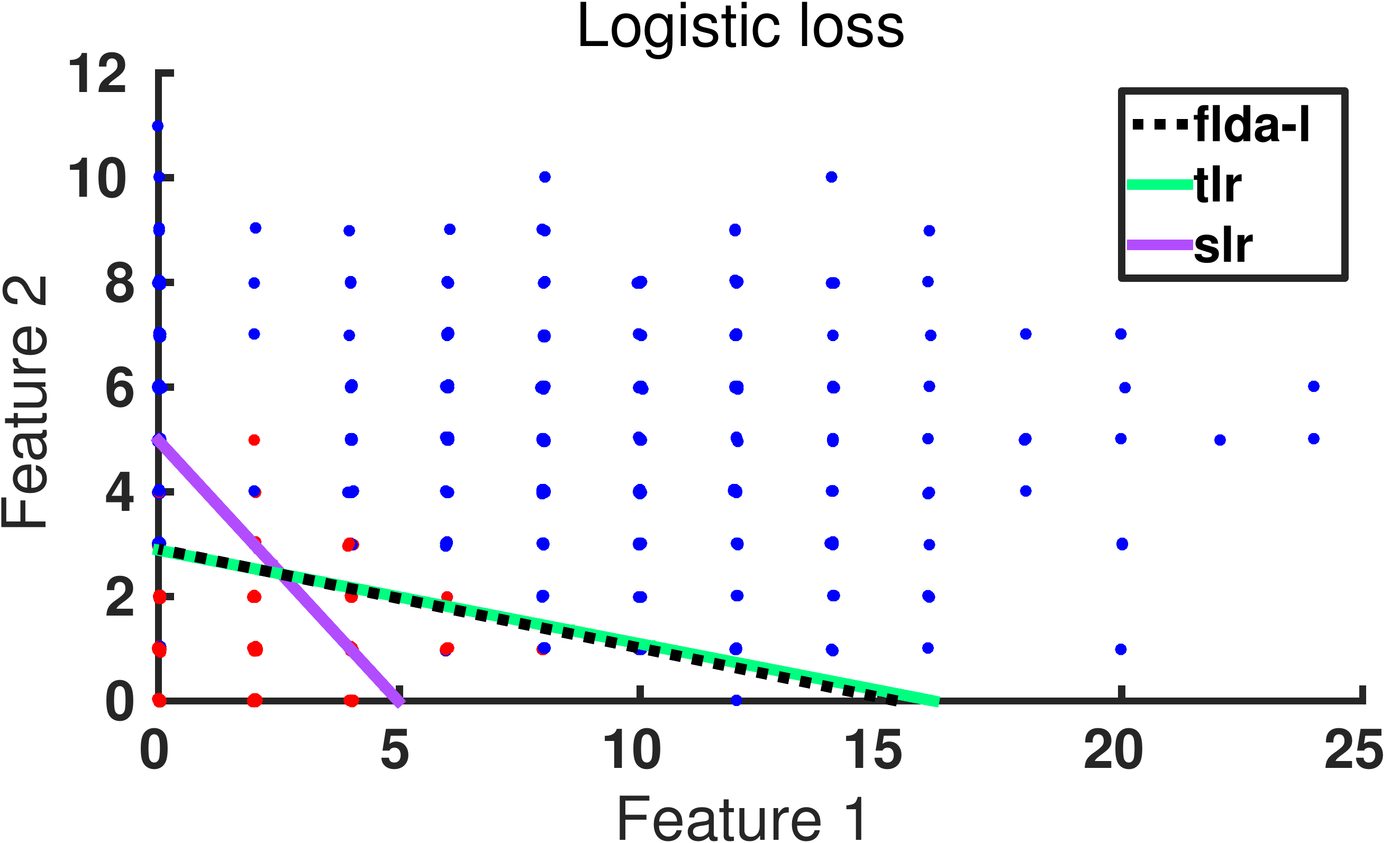}
\caption{Scatter plots of the target domain with decision boundaries of classifiers. The data was generated by bivariate Poisson class-conditional distributions. The decision boundaries were constructed using the source classifier ({\sc s-lr}/{\sc s-ls}), the target classifier ({\sc t-lr}/{\sc t-ls}), and the adapted classifiers (left {\sc flda-q}, right {\sc flda-l}). Note that the decision boundary of {\sc flda} lies on top of the decision boundary of {\sc t-lr}.}
\label{art2}
\end{figure}

In a second experiment, we generate count features by sampling from bivariate Poisson distributions. Herein, we used rate parameters $\lambda = \begin{bmatrix} 2 \ \ 2 \end{bmatrix}$ for the first class and $\lambda = \begin{bmatrix} 6 \ \ 6 \end{bmatrix}$ for the second class. Again, we construct the target domain data by generating new samples and dropping out the values of feature 1 with a probability of 0.5. In this experiment {\sc s-ls} achieves an error of 0.181 and {\sc t-ls}, {\sc flda-q} achieve an error of 0.099, while {\sc s-lr} achieves an error of 0.170 and {\sc t-lr} and {\sc flda-l} achieve an error of 0.084. Figure \ref{art2} shows the decision boundaries of each of these four classifiers. The results show that {\sc flda} has fully adapted to the domain shift and is essentially equivalent to the target classifier for both loss functions.

\subsubsection{Learning curves}
A question that arises from the previous experiments is how many samples {\sc flda} needs to estimate the transfer parameters and adapt to be (nearly) identical to the target classifier. To answer this question, we performed an experiment in which we computed the classification error rate as a function of the number of training samples. The source training and validation data was generated from the same bivariate Poisson distributions as in Figure \ref{art2}. The target training and corresponding validation data was constructed by generating additional source data and dropping out the first feature with a probability of 0.5. Each of the four data sets contained $10,000$ samples. First, we trained a naive least-squares classifier on the source data ({\sc s-ls}) and tested its performance on both the source and target validation sets as a function of the number of source training samples. Second, we trained a naive least-squares classifier on the target training data ({\sc t-ls}) and tested it on the source and target validation sets as a function of the number target training samples. Third, we trained a an adapted classifier ({\sc flda-q}) on equal amounts of labeled source training data and unlabeled target training data and tested it on both the source and target validation sets. The experiment was repeated $50$ times for every sample size to calculate the standard error of the mean.

\begin{figure}[ht]
\centering
\includegraphics[width=.9\textwidth]{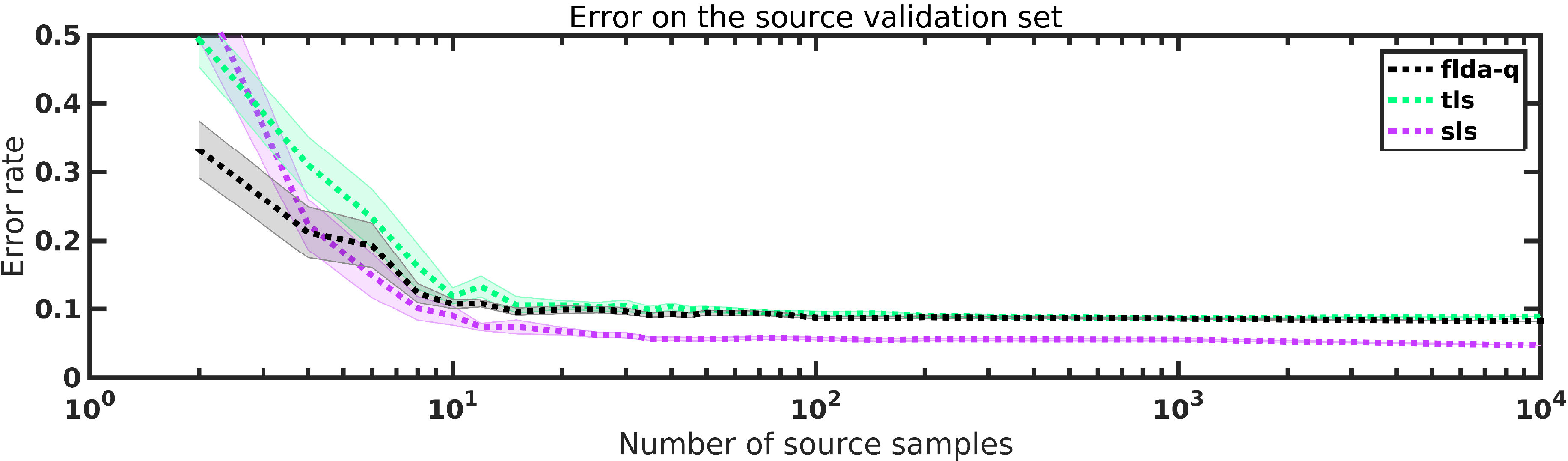}  \vspace{5px} \\
\includegraphics[width=.9\textwidth]{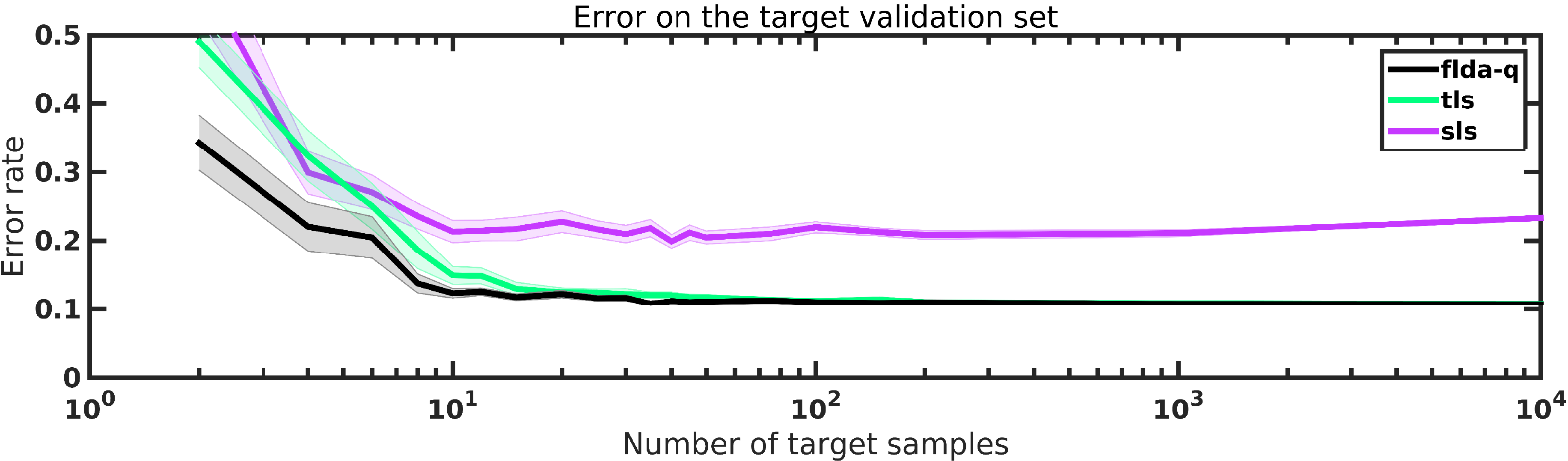}
\caption{Learning curves of the source classifier ({\sc s-ls}), the target classifier ({\sc t-ls}), and adapted classifier ({\sc flda-q}). The top figure shows the error on a validation set generated from two bivariate Poisson distributions. The bottom figure shows the error on a validation set generated from two bivariate Poisson distributions with the first feature dropped out with a probability of 0.5.}
\label{lc}
\end{figure}

The learning curves are plotted in Figure \ref{lc}, which shows the classification error on the source validation set (top) and the classification error on the target validation (bottom). As expected, the source classifier ({\sc s-ls}) outperforms the target ({\sc t-ls}) and adapted ({\sc flda-q}) classifiers on the source domain (dotted lines), while {\sc flda-q} and {\sc t-ls} outperform {\sc s-ls} on the target domain (solid lines). In this problem, it appears that roughly 20 labeled source samples and 20 unlabeled target samples are sufficient for {\sc flda} to adapt to the domain shift. Interestingly, {\sc flda-q} is outperforming {\sc s-ls} and {\sc t-ls} for small sample sizes. This is most likely due to the fact that the application of the transfer model is acting as a kind of regularization. In particular, when the learning curves are computed with $\ell_{2}$-regularized classifiers, then the difference in performance disappears. 

\subsubsection{Parameter estimation errors}
Another question that arises is how sensitive the approach is to estimation errors in the transfer parameters. To answer this question, we performed an experiment in which we artificially introduce an error in the transfer parameters by perturbing them.

As before, we generate $100,000$ samples for both domains by sampling from bivariate Poisson distributions with $\lambda = \begin{bmatrix} 2 \ \ 2 \end{bmatrix}$ for class 1 and $\lambda = \begin{bmatrix} 6 \ \ 6 \end{bmatrix}$ for class 2. Again, the target domain is constructed by dropping out feature 1 with a probability of 0.5. We trained a naive classifier on the source data ({\sc sl}), a naive classifier on the target data ({\sc tl}), and an adapted classifier {\sc flda} with four different sets of parameters: the maximum likelihood estimate of the first transfer parameter $\hat{\theta}_{1}$ plus 0, 0.1, 0.2, and 0.3, respectively. Table \ref{err_sens_params} shows the resulting classification errors, which reveal a relatively small effect of perturbing the estimated transfer parameters: the errors only increase by a few percent in this experiment.

\begin{table}[ht]
\centering
\begin{tabular}{ l | r  r  r  r  r  r  }
 & {\sc sl} & {\sc tl} & $\hat{\theta}_1+0$ & $\hat{\theta}_1+0.1$ & $\hat{\theta}_1+0.2$ & $\hat{\theta}_1+0.3$ \\
\hline \\
Quadratic loss 	& 0.245	& 0.137 	& 0.138	& 0.145 	& 0.149	& 0.150 \\
Logistic loss 		& 0.264 	& 0.139 	& 0.139	& 0.140	& 0.142	& 0.146
\end{tabular}
\caption{Classification errors for a naive source classifier, a naive target classifier, and the adapted classifier with a value of 0, 0.1, 0.2, and 0.3 added to the estimate of the first transfer parameter $\hat{\theta}_{1}$.}
\label{err_sens_params}
\end{table}

\begin{figure}[ht]
	\centering
	\includegraphics[width=.48\textwidth]{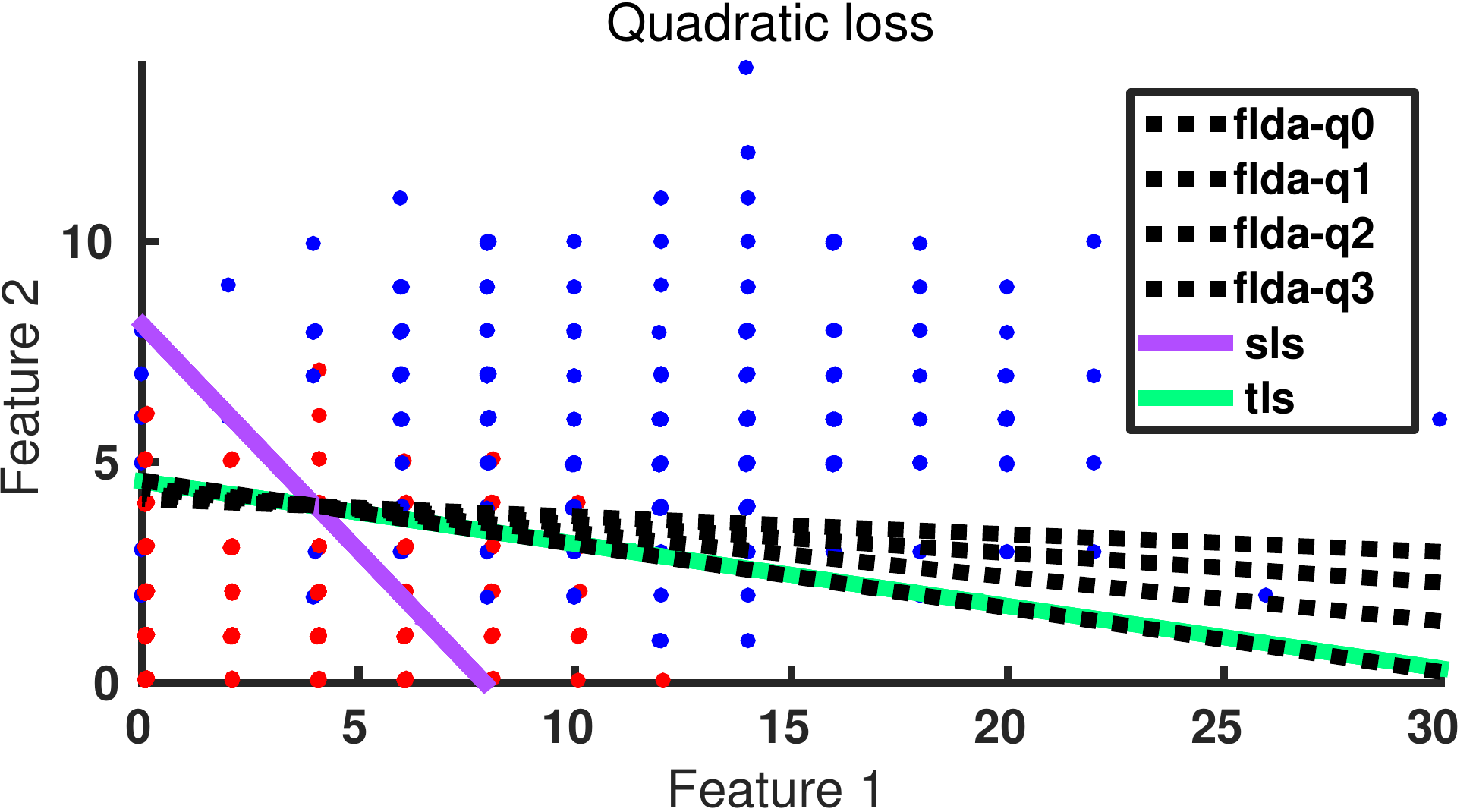} \hfill
	\includegraphics[width=.48\textwidth]{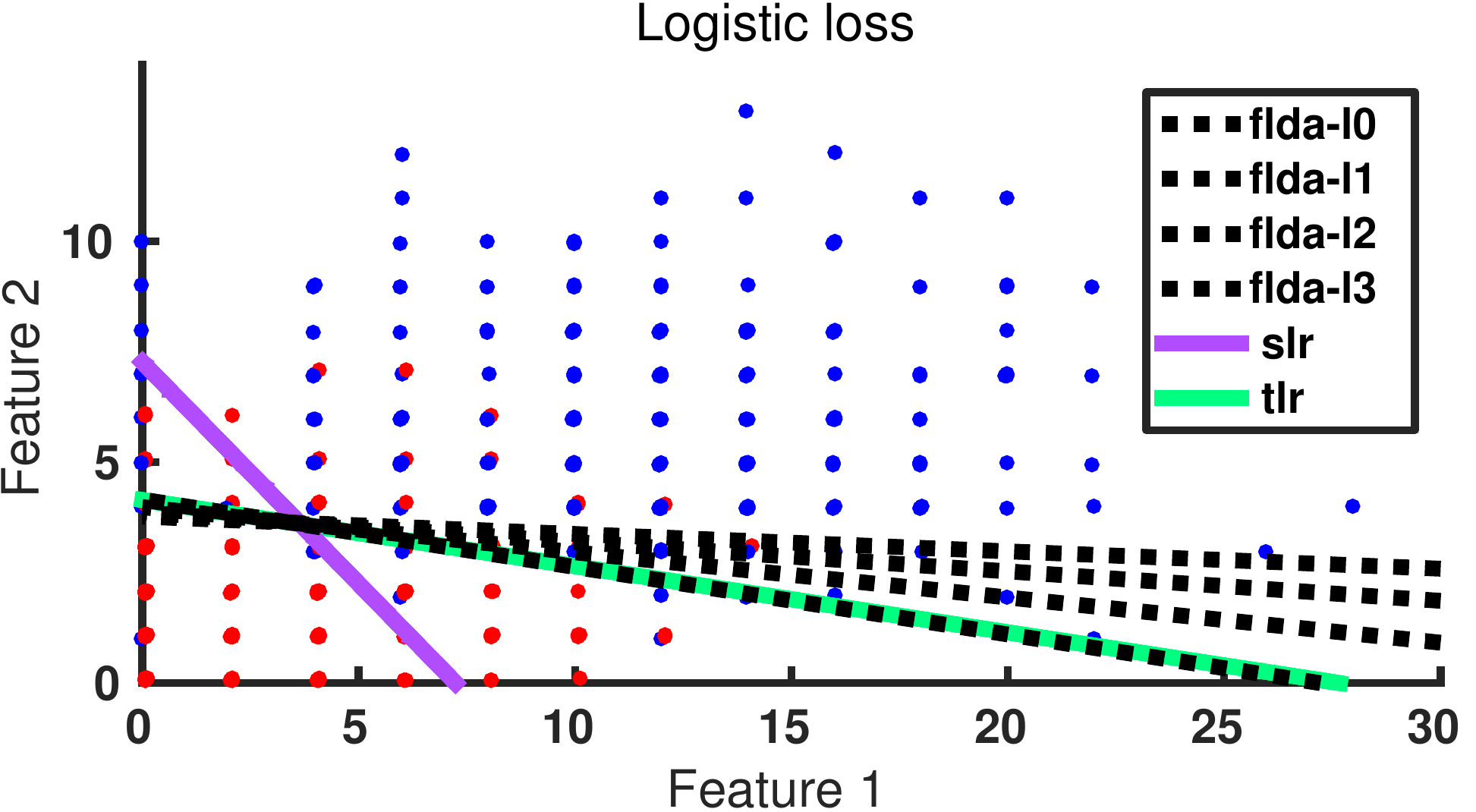} 
	\caption{Scatter plots of the target data and decision boundaries of two naive and four adapted classifiers with transfer parameter estimate errors of 0, 0.1, 0.2, and 0.3. Results are show for both the quadratic loss classifier ({\sc flda-q}; left) and the logistic loss classifier ({\sc flda-l}; right).}
	\label{sens_params}
\end{figure}

Figure \ref{sens_params} (left) shows the decision boundaries for the two naive and four {\sc flda-q} classifiers and (right) shows the two naive and the four {\sc flda-l} classifiers. The figures show that the decision boundary is starting to deviate substantially when the error in the transfer parameter estimates is larger than 0.1. This shows that it is important that the transfer distribution is estimated well for {\sc flda} to produce high-quality classifiers. Having said that, our results do suggest that {\sc flda} is robust to small perturbations in the parameters of the transfer distribution.


\subsection{Natural data}
In the second set of experiments, we evaluate {\sc flda} on a series of real-world datasets and compare it with several state-of-the-art methods. The evaluations are performed in the transductive learning setting that is common on domain adaption: we measure the ability of the classifiers to predict the labels of target samples, using labeled source samples and (unlabeled) target samples during training. 

\subsubsection{Setup}
As baselines, we consider eight alternative methods for domain adaptation. All these methods employ a two-stage procedure. In the first step, the domain adaptation is estimated by finding sample weights, finding domain-invariant features, or finding a transformation of the feature space. In the second step, a classifier is trained using the results of the first stage: the classifier may incorporate reweighed samples during training, add a domain-invariant subspace to the source samples, or transform the source samples according to some transformation. In all experiments, we estimate the hyperparameters, such as $\ell_2$-regularization parameters, via cross-validation on held-out source data. Although this results in optimal values for generalizing to the source domain, it should be noted that these values are not necessarily the optimal values for generalizing to the target domain (\citealp{sugiyama2007covariate}). Each of the eight baseline methods is described briefly below.

\paragraph{Naive Support Vector Machine ({\sc s-svm})} Our first baseline method is a support vector machine trained on only the source samples and applied on the target samples. We made use of the libsvm package by \cite{CC01a} with a radial basis function kernel and we performed cross-validation to estimate the kernel bandwidth and the $\ell_2$-regularization parameter. All multi-class classification is done through an one-vs-one scheme. This method can be readily compared to subspace alignment ({\sc sa}) and transfer component analysis ({\sc tca}) to evaluate the effects of the respective adaptation approaches.

\paragraph{Naive Logistic Regression ({\sc s-lr}).} Our second baseline method is an $\ell_2$-regularized logistic regressor trained on only the source samples. Its main difference with the support vector machine is that it uses a linear model, a logistic loss instead of a hinge loss, and that it has a natural extension to multiclass as opposed to one-vs-one. The value of the regularization parameter was set via cross-validation. This method can be readily compared to kernel mean matching ({\sc kmm}), structural correspondence learning ({\sc scl}), as well as to the logistic loss version of feature-level domain adaptation ({\sc flda-l}).

\paragraph{Kernel Mean Matching ({\sc kmm}).} Kernel mean matching (\citealp{huang2007correcting}) finds importance weights by minimizing the maximum mean discrepancy (MMD) between the reweighed source samples and the target samples. To evaluate the empirical MMD, we used the radial basis function kernel. The weights are then incorporated in an importance weighted $\ell_{2}$-regularized logistic regressor.

\paragraph{Structural Correspondence Learning ({\sc scl}).} In order to build the domain-invariant subspace (\citealp{blitzer2006domain}), the 20 features with the largest proportion of non-zero values in both domains are selected as the pivot features. Their values were dichotomized (1 if $x\neq0$, 0 if $x=0$) and predicted using a modified Huber loss (\citealp{ando2005framework}). The resulting classifier weight matrix was subjected to an eigenvalue decomposition and the eigenvectors with the 15 largest eigenvalues are retained. The source and target samples are both projected onto this basis and the resulting subspaces are added as features to the original source and target feature spaces, respectively. Consequently, classification is done by training an $\ell_2$-regularized logistic regressor on the augmented source samples and testing on the augmented target samples.

\paragraph{Transfer Component Analysis ({\sc tca}).} For transfer component analysis, the closed-form solution to the parametric kernel map described in \cite{pan2011domain} is computed using a radial basis function kernel. Its hyperparameters, \emph{i.e.} kernel bandwidth, the number of dimensions to retain and the trade-off parameter $\mu$, are estimated through cross-validation. After mapping the data onto the transfer components, we trained a support vector machine with an RBF kernel, cross-validating over its kernel bandwidth and the regularization parameter.

\paragraph{Geodesic Flow Kernel ({\sc gfk}).} The geodesic flow kernel is extracted based on the difference in angles between the principal components of the source and target samples (\citealp{gong2012geodesic}). The basis functions of this kernel implicitly map the data onto all possible $d$-dimensional subspaces on the geodesic path between domains. Classification is performed using a kernel 1-nearest neighbor classifier. We used the subspace disagreement measure (SDM) to select an optimal value for the subspace dimensionality $d$.

\paragraph{Subspace Alignment ({\sc sa}).} For subspace alignment (\citealp{fernando2013unsupervised}), all samples are normalized by their sum and all features are z-scored before extracting principal components which are reduced to dimensionality $d$ according to the subspace disagreement measure (SDM) (\citealp{gong2012geodesic}). Subsequently, the Frobenius norm between the transformed source components and target components is minimized with respect to an affine transformation matrix. After projecting the source samples onto the transformed source components, a support vector machine with a radial basis function kernel is trained with cross-validated hyperparameters and tested on the target samples mapped onto the target components.

\paragraph{Target Logistic Regression ({\sc t-lr}).} Finally, we trained a $\ell_2$-regularized logistic regressor using the normally unknown target labels as the oracle solution. This classifier is included to obtain an upper bound on the performance of our classifiers: it measures the performance of a classifier that has access to labeled target samples.

\subsubsection{Missing data at test time}
In this set of experiments, we study ''missing data at test time'' problems in which dropout transfer occurs naturally. Suppose that for the purposes of building a classifier, a dataset is neatly collected with all features measured for all samples. At test time, however, the samples obtained have missing features, for instance, because of sensor failure --- the missing features are replaced by zeros. In this setting, there is a mismatch between the amount of features present in the training data (source domain) and the amount of features present in the test data (target domain). Our approach naturally deals with this lack of information because the missing data can be treated as being {\it dropped out}. We have collected six datasets from the UCI machine learning repository (\citealp{Lichman:2013}) that contain data missing at random: Hepatitis (hepat.), Ozone (ozone; \citealp{zhang2008forecasting}), Heart Disease (heart;  \citealp{detrano1989international}), Mammographic masses (mam.; \citealp{elter2007prediction}), Automobile (auto), and Arrhythmia (arrhy.; \citealp{guvenir1997supervised}). Table \ref{uci-data} shows summary statistics for these sets.

\begin{table}[ht]
\centering
\begin{tabular}{ l | r  r  r  r  r  r  }
 & hepat. & ozone & heart & mam. & auto. & arrhy. \\
\hline \\
Features & 19 & 72 & 13 & 4 & 24 & 279 \\
Samples & 155 & 2534 & 704 & 961& 205 & 452\\
Classes & 2 & 2 & 2 & 2 & 6 & 13  \\
Missing & 75 & 685 & 615 & 130 & 72 & 384
\end{tabular}
\caption{Summary statistics of the UCI repository datasets with missing data.}
\label{uci-data}
\end{table}

In the experiments, we construct the training set (source domain) to contain all samples that do not have any missing values; the test set (target domain) contains the remaining samples, \emph{i.e.} all samples that do have missing values. We replace the missing values by zeros, train the classifiers are trained on the source domain, and evaluate them on the target domain. We note that instead of doing zero-imputation, we also could have used methods such as mean-imputation (\citealp{rubin1976inference, rubin2002statistical}): the {\sc flda} framework naturally allows defining a transfer model that replaces a feature value by its mean instead by a zero value.

\begin{table}[ht]
\centering
\begin{tabular}{ l | c c c c c c c c c | c}
	& {\sc s-svm} & {\sc s-lr} & {\sc kmm} & {\sc scl} & {\sc sa} & {\sc gfk} & {\sc tca} & {\sc flda-q} & {\sc flda-l} & {\sc t-lr} \\
\midrule
hepat. 	& .213		& .493		& .347 	 	& .480		& .253 			& .227 			& .213 			& .227		 	& {\bf .200}	 & .150 		\\
ozone 	& .060		& .124		& .126	 	& .136		& {\bf .047}		& .093			& .140 			& {\bf .047}		 & .079		 & .069		\\
heart 	& .409	 	& .338		& .390	 	& .319		& .596			& .362			& .391 			& {\bf .203}		 & {\bf .203}	 & .177		\\
mam. 	& .331	 	& .462		& .446	 	& .462		& {\bf .323}		& .423			& {\bf .323}		& .462		 	& .431		 & .194		\\
auto. 	& .848	 	& .935 		& .913	 	& .935		& .587			& {\bf .565}		& .848 			& .848		 	& .848		 & .371		\\
arrhy. 	& .930	 	& .854		& .620	 	& .818		& {\bf .414}		& .651			& .930 			& .456		 	& .889		 & .353		
\end{tabular}
\caption{Classification error of ten (domain-adaptation) classifiers on 6 UCI datasets with missing data. All classifier are trained on a source dataset consisting of all observations with no missing data. The classification error was measured on a target dataset constructed by selecting all observations with missing data.}
\label{missing}
\end{table}

Table \ref{missing} reports the classification error rate of all domain-adaptation methods on all datasets. In the table, we bold-faced the lowest error rates for that particular dataset. From the results presented in the table, we observe that whilst there appears to little difference between the domains in the Hepatitis and Ozone datasets, there is substantial domain shift in the other datasets: the naive classifiers even perform at chance level on the Arrhythmia and Automobile datasets. On almost all datasets, both {\sc flda-q} and {\sc flda-l} improve substantially over the {\sc s-lr}, which suggests that they are successfully adapting to the missing data at test time. By contrast, most of the other domain-adaptation techniques do not consistently improve although, admittedly, sample transformation methods appear to work reasonable well on the Ozone, Mammography, and Arrhythmia datasets.

\subsubsection{Handwritten digits}
Handwritten digit datasets have been popular in machine learning because of the large sample size and the interpretability of the images. Generally, the data is acquired by assigning an integer value between 0 and 255 proportional to the amount of pressure that is applied at a particular spatial location on an electronic writing pad. Therefore, the probability of a non-zero value of a pixel informs us how often a pixel is part of a particular digit. For instance, the middle pixel in the digit $8$ is a very important part of the digit because it nearly always corresponds to a high-pressure location, but the upper-left corner pixel is not important. Domain shift may be present between digit datasets due to differences in recording conditions. As a result, we may observe discriminative pixels in one dataset (the source domain) that are hardly ever observed in another dataset (the target domain). As a result, these pixels cannot be used to classify digits in the target domain, and we would like to inform the classifier that it should not assign a large weight to such pixels.

Here, we create a domain adaptation problem setting by considering two handwritten digit sets, namely, the MNIST (\citealp{lecun1998gradient}) and the USPS (\citealp{hull1994database}) datasets. In order to create a common feature space, images from both datasets are resized to 16 by 16 pixels. To reduce the discrepancy between the size of MNIST dataset (which contains $60,000$ examples) and the USPS dataset (which contains $9,298$ examples), we only use $14,000$ samples from the MNIST dataset. The classes are balanced in both datasets. 
\begin{figure}[ht]
	\centering
	\includegraphics[width=.3\textwidth]{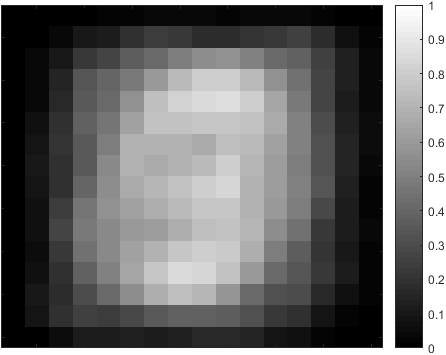} \hspace{40px}
	\includegraphics[width=.3\textwidth]{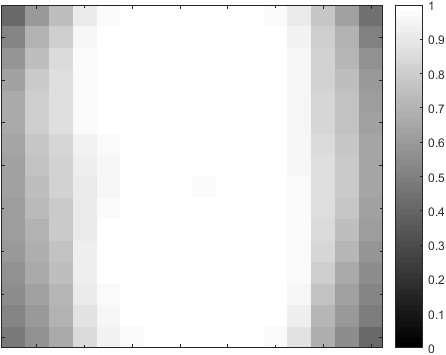}
	\caption{Visualization of the probability of non-zero values for each pixel on the MNIST dataset (left) and the USPS dataset (right).}
	\label{fp_digits}
\end{figure}

\begin{figure}[ht]
	\centering
	\includegraphics[width=.3\textwidth]{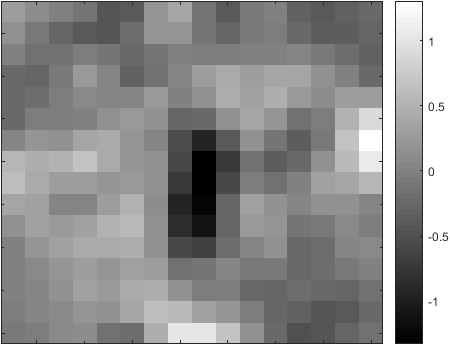} \hfill
	\includegraphics[width=.3\textwidth]{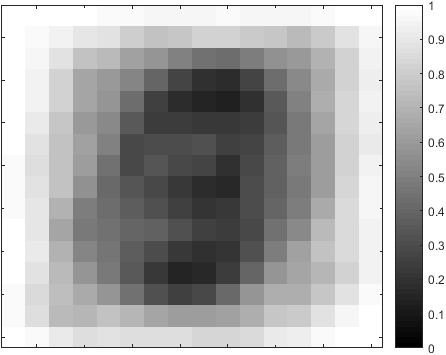} \hfill
	\includegraphics[width=.3\textwidth]{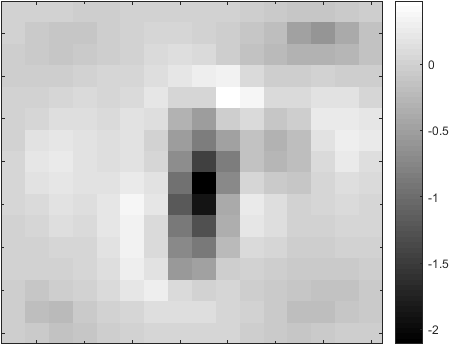}
	\caption{Weights assigned by the naive source classifier to the $0$-digit predictor (left), the transfer parameters of the dropout transfer model (middle), and the weights assigned by the adapted classifier to the $0$-digit predictor for training on USPS images and testing on MNIST ((right; $U \rightarrow M$).}
	\label{tp_digits2}
\end{figure}
Figure \ref{fp_digits} shows a visualization of the probability that each pixel is non-zero for both datasets. The visualization shows that while the digits in the MNIST dataset occupy mostly the center region, the USPS digits tend to occupy a substantially larger part of the image. Figure \ref{tp_digits2} (left) visualizes the weights of the naive linear classifier ({\sc s-lr}), (middle) the dropout probabilities $\theta$, and (right) the adapted classifier's weights ({\sc flda-l}). The middle image shows that dropout weights are large exactly in regions in which USPS pixels are frequent but MNIST pixels are not. The weights of the naive classifier appear to be shaped in a somewhat noisy pattern. The center itself has negative weights, which implies that if those pixels in a new sample have a low intensity, then the image is more likely to be the $0$ digit. By contrast, the weights of the {\sc flda} classifier are smoothed in the periphery, which indicates that the classifier is placing more value on the center pixels and is essentially ignoring the peripheral ones, which is desired when classifying MNIST digits.

Table \ref{err_digits} shows the classification error rates where the rows correspond to both combinations of treating one dataset as the source domain and the other as the target. The results show that there is a large difference between the naive classifiers and classifiers trained on the target data, which indicates that the domains are highly dissimilar. We note that the error rates of the target classifier on the MNIST dataset are higher than usual for this dataset ({\sc t-lr} has an error rate of 0.234): this happens because of the downsampling of the images to 16x16 pixels and because we use fewer samples for training. The results presented in the table highlight an interesting property of {\sc flda} with dropout transfer: {\sc flda} performs well in settings in which the domain transfer can be appropriately modeled by the transfer distribution, namely, in the U$\rightarrow$M setting where pixels that appear in the source domain (USPS) do not appear in the target domain (MNIST). However, this does not work the other way around: the dropout transfer model cannot represent pixels appearing more often in the target domain than in the source domain, which explains the poor performance in the M$\rightarrow$U setting. To work well in that setting, it is presumably necessary to use a richer transfer model with {\sc flda}, for instance, a bit-swap distribution.
\begin{table}[ht]                                                               
\centering                                                                      
\begin{tabular}{ l | c c c c c c c c c | c}
	& {\sc s-svm} & {\sc s-lr} & {\sc kmm} & {\sc scl} & {\sc sa} & {\sc gfk} & {\sc tca} & {\sc flda-q} & {\sc flda-l} & {\sc t-lr} \\
\midrule
M$\rightarrow$U 	& .522	& .747	& .748 	& .747	& .890		& {\bf .497}	& .808		& .811 	 	& .678	 & .055	\\
U$\rightarrow$M 	& .766	& .770	& 769 	& .808	& .757		& .660		& .857		& {\bf .640} 	& .684	 & .234		
\end{tabular}                                                                   
\caption{Classification error rates obtained by ten (domain-adapted) classifiers on both pairs of domains on the handwritten digits data (M='MNIST' and U='USPS'). }    
\label{err_digits}
\end{table} 

\subsubsection{Office-Caltech}
The Office-Caltech dataset (\citealp{hoffman2013efficient}) consists of images of objects gathered using four different methods: one from images found through a web image search (Caltech256), one from images of products on Amazon, one taken with a digital SLR camera and one taken with a webcam. Overall, the set contains 10 classes, with 1123 samples from Caltech, 958 samples from Amazon, 157 samples from the DSLR camera, and 295 samples from the webcam. Our first experiment with the Office-Caltech dataset is based on features extracted through SURF features (\citealp{bay2006surf}). These descriptors determine a set of interest points by finding local maxima in the determinant of the image Hessian. Weighted sums of Haar features are computed in multiple subwindows at various scales around each of the interest points. The resulting descriptors are vector-quantized to produce a bag-of-visual-words histogram of the image that is both scale and rotation-invariant. We perform domain-adaptation experiments by training on one domain and testing on another.

Table \ref{err_office_caltech} shows the results of the classification experiments, where compared to competing methods, {\sc sa} is performing well for a number of domain pairs, which may indicate that the SURF descriptor representation leads to domain dissimilarities that can be accurately captured by subspace transformations. This result is further supported by the fact that the transformations found by {\sc gfk} and {\sc tca} are also outperforming {\sc s-svm}. {\sc flda-q} and {\sc flda-l} are among the best performers on certain domain pairs. In general, {\sc flda} does appear to  perform at least as good or better than a naive {\sc s-lr} classifier.

\begin{table}[ht]                                                               
\centering                                                                      
\begin{tabular}{ l | c c c c c c c c c | c}
	& {\sc s-svm} & {\sc s-lr} & {\sc kmm} & {\sc scl} & {\sc sa} & {\sc gfk} & {\sc tca} & {\sc flda-q} & {\sc flda-l} & {\sc t-lr} \\
\midrule
A$\rightarrow$D & {\bf .599} 	& .618 		& .616 		& .621 		& .627 		& .624 		& .624 	& {\bf .599}	& .624 		& .303 \\
A$\rightarrow$W & .688 		& .675  		& .668 		& .686 		& {\bf .606}	& .631		& .712 	& .648 		& .678 		& .181 \\
A$\rightarrow$C & .557		& .553 		& .563 		& .555 		& .594		& .614 		& .579 	& .565 		& {\bf .550}	& .427 \\
D$\rightarrow$W & .312 		& .312 		& .346 		& .317 		& .167 		& {\bf .153} 	& .295 	& .322 		& .312 		& .181\\
D$\rightarrow$C & .744 		& .712 		& .734 		& .712 		& {\bf .655} 	& .706 		& .680 	& .712 		& .710 		& .427\\
W$\rightarrow$C & .721 		& .698 		& .709 		& .705 		& .677 		& .697 		& .688 	& {\bf .675}	& .701 		& .427\\
D$\rightarrow$A & .876 		& .719 		& .727 		& .724 		& {\bf .616} 	& .680 		& .650 	& .700 		& .722 		& .258\\
W$\rightarrow$A & .676 		& .695 		& .706 		& .707 		& {\bf .631} 	& .665 		& .668 	& .671 		& .691 		& .258\\
C$\rightarrow$A & .493 		& .523 		& .515 		& .496		& .538 		& .592 		& .504 	& .490 		& {\bf .475}	& .258\\
W$\rightarrow$D & .198 		& .191 		& .178 		& .198 		& .214 		& {\bf .121} 	& .166 	& .191 		& .185 		& .303\\
C$\rightarrow$D & .612 		& .616 		& .631 		& .583 		& .575		& .599 		& .612 	& {\bf .510}	& .599 		& .303\\
C$\rightarrow$W & .712 		& .725 		& .729 		& .724 		& {\bf .600}	& .603		& .695 	& .654		& .702 		& .181\\
\end{tabular}                                                                   
\caption{Classification error rates obtained by ten (domain-adapted) classifiers for all pairwise combinations of domains on the Office-Caltech dataset with SURF features (A='Amazon', D='DSLR', W='Webcam', and C='Caltech').}    
\label{err_office_caltech} 
\end{table} 

The results on the Office-Caltech dataset depend on the type of information the SURF descriptors are extracting from the images. We also studied the performance of domain-adaptation methods on a richer visual representation, produced by a pre-trained convolutional neural network. Specifically, we used a dataset provided by \citealp{donahue2014decaf}, who extracted 1000-dimensional feature-layer activations (so-called DeCAF$_{8}$ features) in the upper layers of the a convolutional network that was pre-trained on the Imagenet dataset. \citet{donahue2014decaf} used a larger superset of the Office-Caltech dataset that contains 31 classes with 2817 images from Amazon, 498 from the DSLR camera, and 795 from the webcam. The results of our experiments with the DeCAF$_{8}$ features are presented in Table \ref{err_office_cnn}. The results show substantially lower error rates overall, but they also show that domain transfer in the the DeCAF$_8$ feature representation is not amenable to effective modeling by subspace transformations. {\sc kmm} and {\sc scl} obtain performances that are similar to the of the naive {\sc s-lr} classifier but in one experiment, the naive classifier is actually the best-performing model. Whilst achieving the best performance on 2 out of 6 domain pairs, the {\sc flda-q} and {\sc flda-l} models are not as effective as on other datasets, presumably, because dropout is not a good model for the transfer in a continuous feature space such as the DeCAF$_8$ feature space.
\begin{table}[ht]                                                               
\centering                                                                      
\begin{tabular}{ l | c c c c c c c c c | c}
	& {\sc s-svm} & {\sc s-lr} & {\sc kmm} & {\sc scl} & {\sc sa} & {\sc gfk} & {\sc tca} & {\sc flda-q} & {\sc flda-l} & {\sc t-lr} \\
\midrule
A$\rightarrow$D & .406	& .388		& .402	 	& .422		& .460	& .424		& {\bf .351} 		& .428	 	& .388 	& .104	\\
A$\rightarrow$W & .434	& .468		& .455	 	& .474		& .499	& .477		& {\bf .426}		& .491	 	& .468 		& .064	\\
D$\rightarrow$W & .086	& .079		& .083	 	& .074		& .103	& {\bf .073}	& .087			& .088	 	& .079	 	& .064	\\
D$\rightarrow$A & .516	& .496		& .502	 	& .497		& .520	& .569		& .489			& .589	 	& {\bf .487}	 & .216	\\
W$\rightarrow$A & .520	& {\bf .496}	& .514	 	& .506		& .541	& .584		& .510			& .645	 	& .501	 	& .216	\\
W$\rightarrow$D & .034	& .030		& .032 		& .034		& .062	& .052		& .042			& {\bf .024} 	& .044	 	& .104	
\end{tabular}                                                                   
\caption{Classification error rates obtained by ten (domain-adapted) classifiers for all pairwise combinations of domains on the Office dataset with $\text{DeCAF}_{8}$ features (A='Amazon', D='DSLR', and W='Webcam').}    
\label{err_office_cnn}                                              
\end{table} 

\subsubsection{IMDB}
\begin{figure}[ht]
\centering
\includegraphics[width=.9\textwidth]{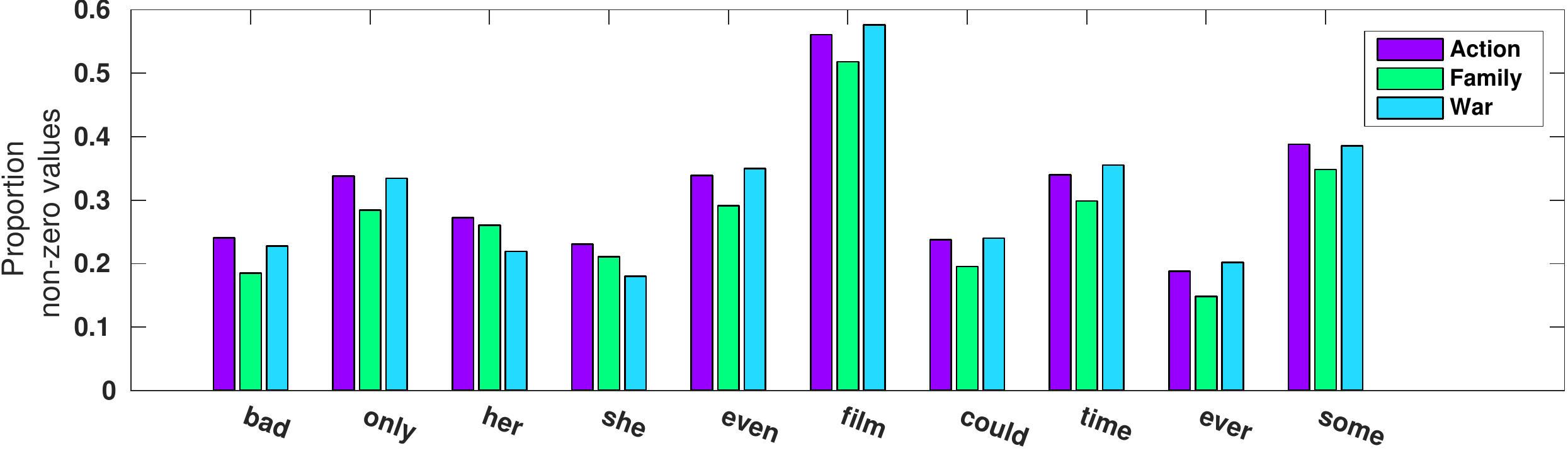}
\caption{Proportion of non-zero values for a subset of words per domain on the IMDB dataset.}
\label{eg_imdb}
\end{figure}

The IMDB movie database (\citealp{pang2004sentimental}) contains written reviews of movies labeled with a 1-10 star rating. The labels are dichotomized with ratings $>$ 5 as +1 and ratings $\leq$ 5 as -1. Using this dichotomy, both classes are roughly balanced. From the original bag-of-words representation, we selected only the features with more than 100 non-zero values in the entire dataset, resulting in 4180 features. To obtain the domains, we split the dataset by genre and obtained 3402 reviews of action movies, 1249 reviews of family movies, and 3697 reviews of war movies. We thus assume that people tend to use different words to review different genres of movies, and we are interested in predicting viewer sentiment after adapting to changes in the word frequencies. To visualize whether this assumption is valid, we plot the proportion of non-zero values of 10 randomly chosen words per domain in Figure \ref{eg_imdb}. The figure suggests that action movie and war movie reviews are quite similar (as expected), but the word use in family movie reviews does appear to be different.

Table \ref{err_imdb} reports the results of the classification experiments on the IMDB database. The first thing to note is that the performances of {\sc s-lr} and {\sc t-lr} are quite similar, which suggests that the frequencies of discriminative words do not vary too much between genres. The results also show that {\sc gfk} and {\sc tca} are not as effective on this dataset as they were on the handwritten digits and Office-Caltech datasets, which suggests that finding a joint subspace that is still discriminative is hard, presumably, because only a small number of the 4180 words actually carry discriminative information. {\sc flda-q} and {\sc flda-l} are better suited for such a scenario, which is reflected by their competitive performance on all domain pairs.

\begin{table}[ht]                                                               
\centering                                                                      
\begin{tabular}{ l | c c c c c c c c c | c}
	& {\sc s-svm} & {\sc s-lr} & {\sc kmm} & {\sc scl} & {\sc sa} & {\sc gfk} & {\sc tca} & {\sc flda-q} & {\sc flda-l} & {\sc t-lr} \\
\midrule
A$\rightarrow$F & .145 & .136 & {\bf .133} & {\bf .133} & .184 & .276 & .230 & .135 & .136 & .196 \\
A$\rightarrow$W & .158 & .155 & .155 & .165 & .163 & .249 & .266 & .158 & {\bf .154} &  .163 \\
F$\rightarrow$W & .256 & .206  & .208 &  .206 & .182 & .289 & .355 & .205 & {\bf .202} &  .163 \\
F$\rightarrow$A & .201 & .195 & {\bf .193} & .198 & {\bf .193} & .296 & .363 & .194 & .194 &  .169 \\
W$\rightarrow$A & .168 & .160 & .159 & .159 & .167 & .238  & .222 & {\bf .155} & .157 & .169  \\
W$\rightarrow$F & .340 &  .167 & .163 &  .163 & .232 & .292 & .203 & .172 & {\bf .159} &  .196
\end{tabular}                                                                   
\caption{Classification error rates obtained by ten (domain-adapted) classifiers for all pairwise combinations of domains on the IMDB dataset. (A='Action', F='Family', and W='War'). }    
\label{err_imdb}                                                      
\end{table} 

\subsubsection{Spam}
\begin{figure}[ht]
	\centering
	\includegraphics[width=.9\textwidth]{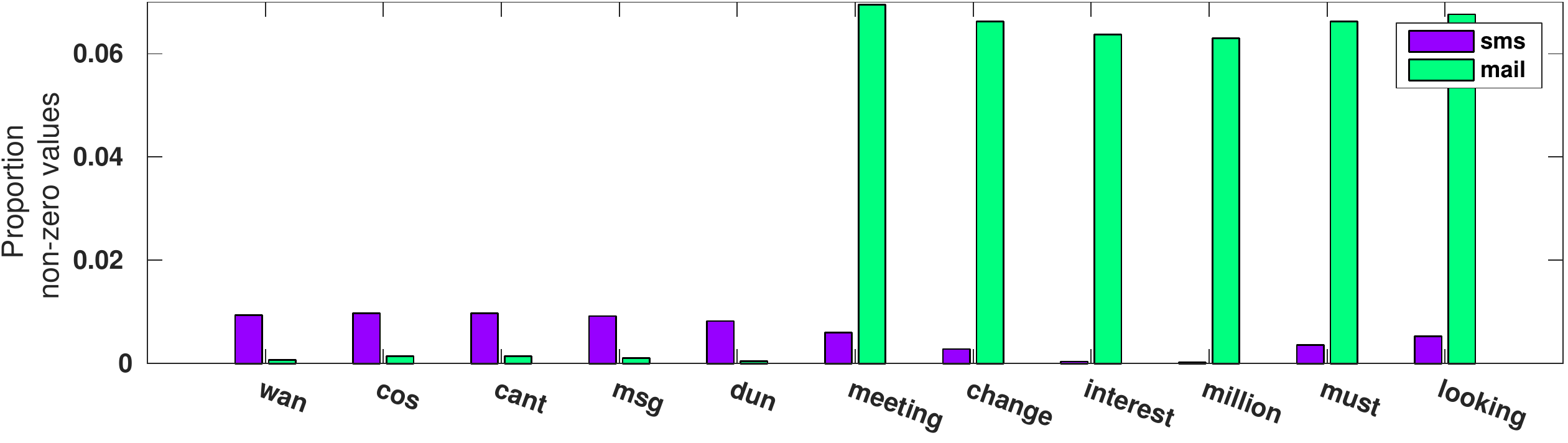}
	\caption{Proportion of non-zero values for a subset of words per domain on the spam dataset.}
	\label{eg_spam}
\end{figure}

Domain adaptation problems may also arise when developing spam detection systems. In our spam-detection experiment, we collected two datasets from the UCI machine learning repository: one containing 4205 emails from the Enron spam database (\citealp{klimt2004introducing}) and one containing 5338 text messages from the SMS-spam dataset (\citealp{almeida2011contributions}). Both were represented using bag-of-words vectors over 4272 words that occurred in both datasets. Figure \ref{eg_spam} shows the proportions of non-zero values for some example words, and shows that there exist large differences in word frequencies between the two domains. In particular, much of the domain differences are due to text messages using shortened words, whereas email messages tend to be more formal. 

Table \ref{err_spam} shows results from our classification experiments on the spam dataset. As can be seen from the results of {\sc t-lr}, fairly good accuracies can be obtained on the spam detection task. However, the domains are so different that the naive classifiers {\sc s-svm} and {\sc s-lr} are performing according to chance or worse. Most of the domain-adaptation models do not appear to improve much over the naive models. For {\sc kmm} this makes sense, as the importance weight estimator will assign equal values to each sample when the empirical supports of the two domains are disjoint. There might be some features that are shared between domains, \emph{i.e.}, words that are spam in both emails and text messages, but considering the performance of {\sc scl} these might not be corresponding well with the other features. {\sc flda-q} and {\sc flda-l} are showing slight improvements over the naive classifiers, but the transfer model we used is apparently too simple, in particular, because the dropout distribution is not modeling the increased frequencies of some words in the other domain.
\begin{table}[ht]                                                               
\centering                                                                      
\begin{tabular}{ l | c c c c c c c c c | c}
	& {\sc s-svm} & {\sc s-lr} & {\sc kmm} & {\sc scl} & {\sc sa} & {\sc gfk} & {\sc tca} & {\sc flda-q} & {\sc flda-l} & {\sc t-lr} \\
\midrule
S$\rightarrow$M & .460	& .522	& .521 	 	& .524		& {\bf .445}	& .491		& .508		& .511	 & .521	 & .073	\\
M$\rightarrow$S & .830	& .804	& .799	 	& .804		& {\bf .408}	& .696		& .863		& .636	 & .727	 & .133	
\end{tabular}                                                                   
\caption{Classification error rates obtained by ten (domain-adapted) classifiers for both domain pairs on the spam dataset. (S='SMS' and M='E-Mail').}    
\label{err_spam}                                                      
\end{table} 

\subsubsection{Amazon}
\begin{figure}[ht]
\centering
\includegraphics[width=.9\textwidth]{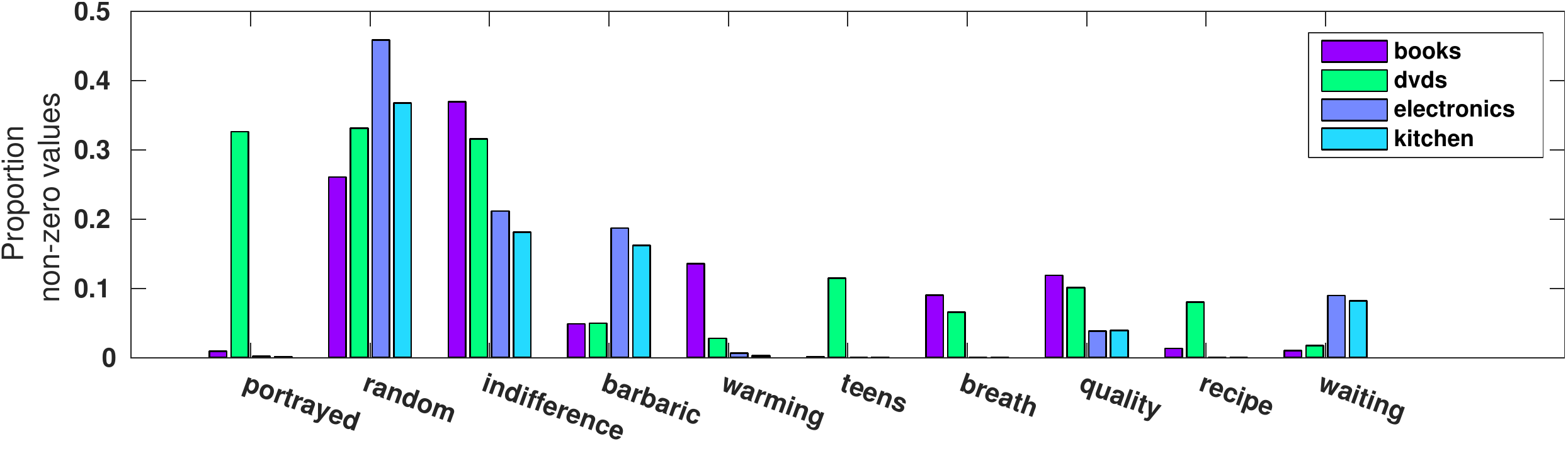}
\caption{Proportion of non-zero values for a subset of words per domain on the Amazon dataset.}
\label{eg_amazon}
\end{figure}

We performed a similar experiment on the Amazon sentiment analysis dataset of product reviews by \cite{blitzer2007biographies}. The data is contains $30,000$ dimensional bag-of-words representations of $27,677$ reviews with the labels derived from the dichotomized 5-star rating (ratings $>$ 3 are +1 and ratings $\leq$ 3 as -1). Each review describes a product from one of four categories: books (6465 reviews), DVDs (5586 reviews), electronics (7681 reviews) and kitchen appliances (7945 reviews). Figure \ref{eg_amazon} shows the probability of a non-zero value for some example words in each category. Some words, such as 'portrayed' or 'barbaric', are very specific to one or two domains, but the frequencies of many other words do not vary much between domains. We performed experiments on the Amazon dataset using the same experimental setup as before.

In Table \ref{err_amazon}, we report the classification error rates on all pairwise combinations of domains. The difference in classification errors between {\sc s-lr} and {\sc t-lr} is up to 10\%, which suggests there is potential for success with domain adaptation. However, the transfer between the domains that is not captured well by {\sc sa}, {\sc gfk}, {\sc tca}: on average, these methods are performing worse than the naive classifiers. We presume this happens because only a small number of words are actually discriminative, and these words carry little weight in the sample transformation measures used. Furthermore, there are significantly less samples than features in each domain which means models with large amounts of parameters are likely to experience estimation errors. By contrast, {\sc flda-l} performs strongly on the Amazon dataset, achieving the best performance on many of the domain pairs. {\sc flda-q} performs substantially worse than {\sc flda-l}, presumably, because of the singular covariance matrix and the fact that quadratic losses are very sensitive to outliers in the labels.
\begin{table}[ht]                                                               
\centering                                                                      
\begin{tabular}{ l | c c c c c c c c c | c}
	& {\sc s-svm} & {\sc s-lr} & {\sc kmm} & {\sc scl} & {\sc sa} & {\sc gfk} & {\sc tca} & {\sc flda-q} & {\sc flda-l} & {\sc t-lr} \\
\midrule
B$\rightarrow$D & .180		& .168		& {\bf .166}	& .167		& .414	& .392		& .413	& .303	& {\bf .166}	& .153	\\
B$\rightarrow$E & .217		& .221		& .222	 	& .220		& .372	& .429		& .369	& .343 	& {\bf .210} 	& .116	\\
B$\rightarrow$K & .188		& .188		& .189	 	& {\bf .184}	& .371	& .443		& .338	& .384 	& .185	 	& .095	\\
D$\rightarrow$E & .201		& .202		& .205	 	& .207 		& .403	& .480		& .385	& .369 	& {\bf .196} 	& .116	\\
D$\rightarrow$K & {\bf .182}	& {\bf .182}	& .185	 	& .190		& .330	& .494		& .360	& .379 	& .185	 	& .095	\\
E$\rightarrow$K & .108		& .110		& .106 		& .112 		& .311	& .416		& .261	& .308	& {\bf .104}	 & .095	\\
D$\rightarrow$B & .192		& .190		& .191	 	& .202		& .351	& .388		& .420	& .368 	& {\bf .186}	 & .145	\\
E$\rightarrow$B & .257		& .262		& {\bf .253}	& .260 		& .372	& .445		& .481	& .406 	& .261 		& .145	\\
K$\rightarrow$B & {\bf .261}	& .277		& .268 		& .273		& .414	& .418		& .426	& .399 	& .271 		& .145	\\
E$\rightarrow$D & .245		& .240		& {\bf .238}	&  .242 		& .398	& .441		& .427	& .384 	& {\bf .238} 	& .153	\\
K$\rightarrow$D & .230		& .230		& .230	 	& .231		& .383	& .410		& .400	& .370 	& {\bf .228} 	& .153	\\
K$\rightarrow$E & .123		& .131		& .126 		& .126 		& .290	& .353		& .296	& .292 	& {\bf .119}	 & .116	
\end{tabular}                                                                   
\caption{Classification error rates obtained by ten (domain-adapted) classifiers for all pairwise combinations of domains on the Amazon dataset. (B='Books', D='DVD', E='Electronics', and K='Kitchen').}    
\label{err_amazon}                                                      
\end{table}

\section{Discussion and Conclusions}
We have presented an approach to domain adaptation, called {\sc flda}, that fits a probabilistic model to capture the transfer between the source and the target data and, subsequently, trains a classifier by minimizing the expected loss on the source data under this transfer model. Whilst the {\sc flda} approach is very general, in this paper, we have focused on one particular transfer model, namely, a dropout model. Our extensive experimental evaluation with this transfer model shows that {\sc flda} performs on par with the current state-of-the-art methods for domain adaptation.

An interesting interpretation of our formulation is that the expected loss under the transfer model performs a kind of data-dependent regularization (\citealp{wager2013dropout}). For instance, if a quadratic loss function is employed in combination with the dropout transfer model, {\sc flda} reduces to a transfer-dependent variant of ridge regression (\citealp{bishop1995training}). 
This transfer-dependent regularizer increases the amount of regularization on features if it is undesired for the classifier to assign a large weight to that feature, because the feature is frequently present in the source domain but very infrequently present in the target domain. 
By strongly regularizing the weights corresponding to these features, {\sc flda} achieves the desired goal of essentially ignoring such features in the classifier.


In some of our experiments, the adaptation strategies are producing classifiers that perform worse than a naive classifier trained on the source data. A potential reason for this is that many domain-adaptation models make strong assumptions on the data that may be invalid in many real-world scenarios. In particular, it is unclear to what extent the relation between source data and labels truly is informative about the labels of the target data.
This issue arises in every domain-adaptation problem: without target labels, there is no way of knowing whether matching the target distribution $p_{{\cal Z}}$ to match the source distribution $p_{{\cal X}}$ will improve the match between $p_{{\cal Y} \mid {\cal X}}$ and $p_{{\cal Y} \mid {\cal Z}}$.

\acks{This work was supported by the Netherlands Organization for Scientific Research (NWO; grant 612.001.301). The authors would like to thank Sinno Pan and Boqing Gong for insightful discussions.}

\newpage
\appendix
\section*{Appendix A}
For some combinations of source and target models, the source domain can be integrated out. For others, we would have to resort to Markov Chain Monte Carlo sampling. For the Bernoulli and dropout distributions defined in Equations \ref{sourceB} and \ref{dropout}, the integration as in Equation \ref{z_model} can be performed by plugging in the specified probabilities and performing the summation:
\begin{align}
	q_{{\cal Z}}({\bf z} \mid \eta, \theta) =& \prod_{d=1}^{m} \ \int_{{\cal X}} p_{{\cal Z} \mid {\cal X}}(z_{-d} \mid x_{-d}, \theta_{d}) \ p_{{\cal X}}(x_{-d} \mid \eta_{d}) \ \mathrm{d}x_{-d} \nonumber \\
	=& \prod_{d=1}^{m} \ \sum_{\mathds{1}_{x_{-d} \neq 0}=0}^{1} p_{{\cal Z} \mid {\cal X}}(z_{-d} \mid \mathds{1}_{x_{-d} \neq 0}, \theta_{d}) \ p_{X}(\mathds{1}_{x_{-d} \neq 0}; \eta_{d})  \nonumber \\
	=& \prod_{d=1}^{m} \ \begin{cases} \sum_{\mathds{1}_{x_{-d} \neq 0}=0}^{1} p_{{\cal Z} \mid {\cal X}}(z_{-d} =0 \mid \mathds{1}_{x_{-d} \neq 0}, \theta_{d}) \ p_{{\cal X}}(\mathds{1}_{x_{-d} \neq 0}; \eta_{d}) \quad & \text{if} \ z_{-d} = 0 \\
	\sum_{\mathds{1}_{x_{-d} \neq 0}=0}^{1} p_{{\cal Z} \mid \cal{X}}(z_{-d} \neq 0 \mid \mathds{1}_{x_{-d} \neq 0}, \theta_{d}) \ p_{{\cal X}}(\mathds{1}_{x_{-d} \neq 0}; \eta_{d}) \quad & \text{if} \ z_{-d} \neq 0
	\end{cases} \nonumber \\
	=& \prod_{d=1}^{m} \ \begin{cases} 1(1-\eta_{d}) + \theta_{d}\eta_{d} \quad & \text{if} \ z_{-d} = 0 \\
	0(1-\eta_{d}) + (1-\theta_{d})\eta_{d} \quad & \text{if} \ z_{-d} \neq 0
	\end{cases} \nonumber \\
	=& \ \prod_{d=1}^{m} \ \Big((1-\theta_{d}) \ \eta_{d}\Big)^{\mathds{1}_{{z}_{-d} \neq 0}} \ \Big(1-(1-\theta_{d}) \ \eta_{d}\Big)^{1-\mathds{1}_{z_{-d} \neq 0}} \nonumber
\end{align}
where the subscript of $x_{-d}$ refers to the $d$-th feature of any sample $x_{id}$. Note that we chose our transfer model such that the probability is 0 for a non-zero target sample value given a zero source sample value; $p_{{\cal Z} \mid {\cal X}}(z_{-d} \neq 0 \mid \mathds{1}_{x_{-d} \neq 0}=0, \theta_{d}) = 0$. In other words, if a word is not used in the source domain, then we expect that it is also not used in the target domain. By setting different values for these probabilities, we are modeling different types of transfer.

\section*{Appendix B}
The gradient to the second-order Taylor approximation of binary {\sc flda-l} for a general transfer model is:
\begin{flalign}
\frac{\partial }{\partial {\bf w}}\hat{R}(h \mid S) =& \frac{1}{|S|} \sum_{({\bf{x}_i}, y_i) \in S} -y_{i}\mathbb{E}_{{\cal Z}\mid {\bf x}_i}[{\bf z}] + \frac{\sum_{y' \in Y} y'\exp(y' {\bf w}^{\top}{\bf x}_{i})}{\sum_{y'' \in Y}\exp(y'' {\bf w}^{\top}{\bf x}_{i})}{\bf x}_{i} \ + \nonumber \\
& \Bigg[\Big(1-\big[\frac{\sum_{y' \in Y}y'\exp(y' {\bf w}^{\top}{\bf x}_{i})}{\sum_{y'' \in Y}\exp(y'' {\bf w}^{\top}{\bf x}_{i})}\big]^{2}\Big) {\bf w}^{\top}{\bf x}_{i} + \frac{\sum_{y' \in Y} y'\exp(y' {\bf w}^{\top}{\bf x}_{i})}{\sum_{y'' \in Y}\exp(y'' {\bf w}^{\top}{\bf x}_{i})}\Bigg]\big(\mathbb{E}_{{\cal Z} \mid {\bf x}_i}[{\bf z}]-{\bf x}_{i}\big)  \nonumber \\ 
&+ 4\sigma \big(-2 \ {\bf w}^{\top}{\bf x}_{i}\big)\sigma \big(2 \ {\bf w}^{\top}{\bf x}_{i}\big)  \Bigg[\Big(\big[\sigma(-2 \ {\bf w}^{\top}{\bf x}_{i}) - \sigma(2 \ {\bf w}^{\top}{\bf x}_{i})\big]{\bf w}^{\top}{\bf x}_{i} + 1\Big) \nonumber \\
& {\bf w}^{\top}\Big(\mathbb{V}_{{\cal Z}\mid {\bf x}_i}[{\bf z}] + (\mathbb{E}_{{\cal Z} \mid {\bf x}_i}[{\bf z}]-{\bf x}_{i})(\mathbb{E}_{{\cal Z} \mid {\bf x}_i}[{\bf z}]-{\bf x}_{i})^{\top} \Big) {\bf w}\Bigg] \nonumber
\end{flalign}

\section*{Appendix C}
The second-order Taylor approximation to the expectation over the log-partition function for a multi-class classifier weight matrix ${\bf W}$ of size $(m+1) \times K$ around the point $a_i = {\bf W}^{\top}{\bf x}_i$ is:
\begin{align}
\mathbb{E}_{{\cal Z} \mid {\bf x}_i}\big[A({\bf W}^{\top}{\bf z})\big] \approx& \ A(a_{i}) + A'(a_{i})(\mathbb{E}_{{\cal Z} \mid {\bf x}_i} [{\bf W}^{\top}{\bf z}] - a_{i}) + \frac{1}{2}A''(a_{i})(\mathbb{E}_{{\cal Z} \mid {\bf x}_i} [{\bf W}^{\top}{\bf z}] - a_{i})^{2} \nonumber \\
=& \log \sum_{k=1}^{K} \exp( {\bf W}_{k}^{\top}{\bf x}_{i}) + \sum_{k=1}^{K}\frac{\exp({\bf W}_{k}^{\top}{\bf x}_{i})}{\sum_{k=1}^{K}\exp({\bf W}_{k}^{\top}{\bf x}_{i})}{\bf W}_{k}^{\top} (\mathbb{E}_{{\cal Z} \mid {\bf x}_i}[{\bf z}] - {\bf x}_{i})  \nonumber \\
& +\frac{1}{2} \sum_{k=1}^{K}\left(\frac{\exp({\bf W}_{k}^{\top}{\bf x}_{i})}{\sum_{k=1}^{K}\exp({\bf W}_{k}^{\top}{\bf x}_{i})} - \frac{\exp(2{\bf W}_{k}^{\top}{\bf x}_{i})}{\big(\sum_{k=1}^{K}\exp({\bf W}_{k}^{\top}{\bf x}_{i})\big)^2} \right) \nonumber \\
&{\bf W}_{k}^{\top}\Big(\mathbb{V}_{{\cal Z} \mid {\bf x}_i}[{\bf z}] + (\mathbb{E}_{{\cal Z} \mid {\bf x}_i}[{\bf z}]-{\bf x}_{i})(\mathbb{E}_{{\cal Z} \mid {\bf x}_i}[{\bf z}] - {\bf x}_{i})^{\top} \Big){\bf W}_{k} \nonumber \, .
\label{log_loss}
\end{align}

The results contains a number of recurring terms which means it can be efficiently implemented. Incorporating the multiclass approximation into the loss, we can derive the following gradient:
\begin{align}
\frac{\partial }{\partial {\bf W}_{k}}&\hat{R}(h \mid S) = \frac{1}{|S|}  \sum_{({\bf{x}_i}, y_i) \in S} -y_{i}\mathbb{E}_{{\cal Z} \mid {\bf x}_i}[{\bf z}]  \nonumber \\
& + \left[\frac{\exp({\bf W}_{k}^{\top}{\bf x}_{i})}{\sum_{k}^{K}\exp({\bf W}_{k}^{\top}{\bf x}_{i})} - \frac{\exp(2{\bf W}_{k}^{\top}{\bf x}_{i})}{\big(\sum_{k}^{K}\exp({\bf W}_{k}^{\top}{\bf x}_{i})\big)^2} \right] {\bf x}_{i}{\bf W}_{k}^{\top} (\mathbb{E}_{{\cal Z} \mid {\bf x}_i}[{\bf z}] - {\bf x}_{i}) \nonumber \\
& + \frac{\exp({\bf W}_{k}^{\top}{\bf x}_{i}){\bf x}_{i}}{\sum_{k}^{K}\exp({\bf W}_{k}^{\top}{\bf x}_{i})} (\mathbb{E}_{{\cal Z} \mid {\bf x}_i}[{\bf z}] - {\bf x}_{i}) \nonumber \\
& + \left[ \frac{\exp({\bf W}_{k}^{\top}{\bf x}_{i}){\bf x}_{i}}{\sum_{k}^{K}\exp({\bf W}_{k}^{\top}{\bf x}_{i})} -3 \ \frac{\exp(2{\bf W}_{k}^{\top}{\bf x}_{i}){\bf x}_{i}}{\big(\sum_{k}^{K}\exp({\bf W}_{k}^{\top}{\bf x}_{i})\big)^2} +2 \ \frac{\exp(3{\bf W}_{k}^{\top}{\bf x}_{i}){\bf x}_{i}}{\big(\sum_{k}^{K}\exp({\bf W}_{k}^{\top}{\bf x}_{i})\big)^3} \right] \nonumber \\
& {\bf W}_{k}^{\top}\Big(\mathbb{V}_{{\cal Z} \mid {\bf x}_i}[{\bf z}] + (\mathbb{E}_{{\cal Z} \mid {\bf x}_i}[{\bf z}]-{\bf x}_{i})(\mathbb{E}_{{\cal Z} \mid {\bf x}_i}[{\bf z}] - {\bf x}_{i})^{\top} \Big){\bf W}_{k} \nonumber \\  
& + 2\left[\frac{\exp({\bf W}_{k}^{\top}{\bf x}_{i})}{\sum_{k}^{K}\exp({\bf W}_{k}^{\top}{\bf x}_{i})} - \frac{\exp(2{\bf W}_{k}^{\top}{\bf x}_{i})}{\sum_{k}^{K}\exp({\bf W}_{k}^{\top}{\bf x}_{i})^2} \right] \Big(\mathbb{V}_{{\cal Z} \mid {\bf x}_i}[{\bf z}] + (\mathbb{E}_{{\cal Z} \mid {\bf x}_i}[{\bf z}]-{\bf x}_{i})(\mathbb{E}_{{\cal Z} \mid {\bf x}_i}[{\bf z}] - {\bf x}_{i})^{\top} \Big){\bf W}_{k} \nonumber \\  
& -\frac{\exp({\bf W}_{k}^{\top}{\bf x}_{i}){\bf x}_{i}}{\big(\sum_{k}^{K}\exp({\bf W}_{k}^{\top}{\bf x}_{i})\big)^2} \sum_{j\neq k}^{K}\exp({\bf W}_{j}^{\top}{\bf x}_{i}){\bf W}_{j}^{\top}\Big(\mathbb{V}_{{\cal Z} \mid {\bf x}_i}[{\bf z}] + (\mathbb{E}_{{\cal Z} \mid {\bf x}_i}[{\bf z}]-{\bf x}_{i})(\mathbb{E}_{{\cal Z} \mid {\bf x}_i}[{\bf z}] - {\bf x}_{i})^{\top} \Big){\bf W}_{j} \nonumber \\  
& +2 \ \frac{\exp({\bf W}_{k}^{\top}{\bf x}_{i}){\bf x}_{i}}{\big(\sum_{k}^{K}\exp({\bf W}_{k}^{\top}{\bf x}_{i})\big)^3} \sum_{j\neq k}^{K}\exp(2{\bf W}_{j}^{\top}{\bf x}_{i}){\bf W}_{j}^{\top}\Big(\mathbb{V}_{{\cal Z} \mid {\bf x}_i}[{\bf z}] + (\mathbb{E}_{{\cal Z} \mid {\bf x}_i}[{\bf z}]-{\bf x}_{i})(\mathbb{E}_{{\cal Z} \mid {\bf x}_i}[{\bf z}] - {\bf x}_{i})^{\top} \Big){\bf W}_{j} \nonumber \, .
\end{align}

\newpage
\vskip 0.2in
\bibliography{kouw15a}

\end{document}